\documentclass[letterpaper, 10 pt, journal, twoside]{IEEEtran}
\usepackage{etoolbox}
\usepackage{subfig}
\usepackage{times}
\usepackage{blindtext}
\usepackage{cite}
\usepackage{amsmath,amssymb,amsfonts}
\usepackage{graphicx}
\usepackage{units}
\usepackage{algorithm}
\usepackage[noend]{algpseudocode}
\usepackage{dblfloatfix}    
\usepackage{graphicx, xfrac}
\usepackage{textcomp}
\usepackage{xcolor}
\makeatletter
\let\NAT@parse\undefined
\makeatother
\usepackage[colorlinks]{hyperref}
\hypersetup{citecolor = black, linkcolor = black, urlcolor=black}
\usepackage[T1]{fontenc}
\usepackage{orcidlink}
\usepackage{parnotes}
\DeclareMathOperator{\arctantwo}{arctan2}
\usepackage{array}

\ifCLASSINFOpdf
\else
\fi
\begin{document}
%
\title{UWB Radar SLAM: an Anchorless Approach in Vision Denied Indoor Environments}
%
%
%

\author{H. A. G. C. Premachandra$^{1}$, Ran Liu$^{1}$, Chau Yuen$^{2}$, and U-Xuan Tan$^{1}$%
\thanks{Manuscript received: February, 27, 2023; Revised April, 26, 2023; Accepted June, 23, 2023.}
\thanks{This paper was recommended for publication by Editor Javier Civera upon evaluation of the Associate Editor and Reviewers' comments.
} 
\thanks{$^{1}$The authors are with Singapore University of Technology and Design, Singapore. {\tt\footnotesize gihan\_appuhamilage@mymail.sutd.edu.sg}

$^{2}$The author is with Nanyang Technological University, Singapore.
        }%
\thanks{Digital Object Identifier (DOI): see top of this page.}
}
\maketitle

\begin{abstract}
LiDAR and cameras are frequently used as sensors for simultaneous localization and mapping (SLAM). However, these sensors are prone to failure under low visibility (e.g. smoke) or places with reflective surfaces (e.g. mirrors). \textcolor{black}{On the other hand, electromagnetic waves exhibit better penetration properties when the wavelength increases, thus are not affected by low visibility.} Hence, this paper presents ultra-wideband (UWB) radar as an alternative to the existing sensors. UWB is generally known to be used in anchor-tag SLAM systems. One or more anchors are installed in the environment and the tags are attached to the robots. Although this method performs well under low visibility, modifying the existing infrastructure is not always feasible. UWB has also been used in peer-to-peer ranging collaborative SLAM systems. However, this requires more than a single robot and does not include mapping in the mentioned environment like smoke. Therefore, the presented approach in this paper solely depends on the UWB transceivers mounted on-board. In addition, \textcolor{black}{an} extended Kalman filter (EKF) SLAM is used to solve the SLAM problem \textcolor{black}{at the back-end.} Experiments were conducted and demonstrated that the proposed UWB-based radar SLAM is able to map natural point landmarks inside an indoor environment while improving robot localization. 
\end{abstract}

\begin{IEEEkeywords}
Range Sensing, SLAM.
\end{IEEEkeywords}

\captionsetup[figure]{font=small}
\captionsetup[table]{justification=centering, singlelinecheck=false, labelsep=newline, font={sc,footnotesize}}

%
\IEEEpeerreviewmaketitle

\section{Introduction}\label{intro}
%
%
%
%
\IEEEPARstart{S}{imultaneous} localization and mapping (SLAM) is essential for a mobile robot to navigate through a previously unexplored environment to accomplish a given task. SLAM algorithms usually rely on the sensors attached to the robot. Hence, it is of interest to explore new sensing technologies to overcome the limitations of the existing methods.

\begin{figure}[t]
	\centering
	\subfloat[][]{
		\includegraphics[height=0.95in]{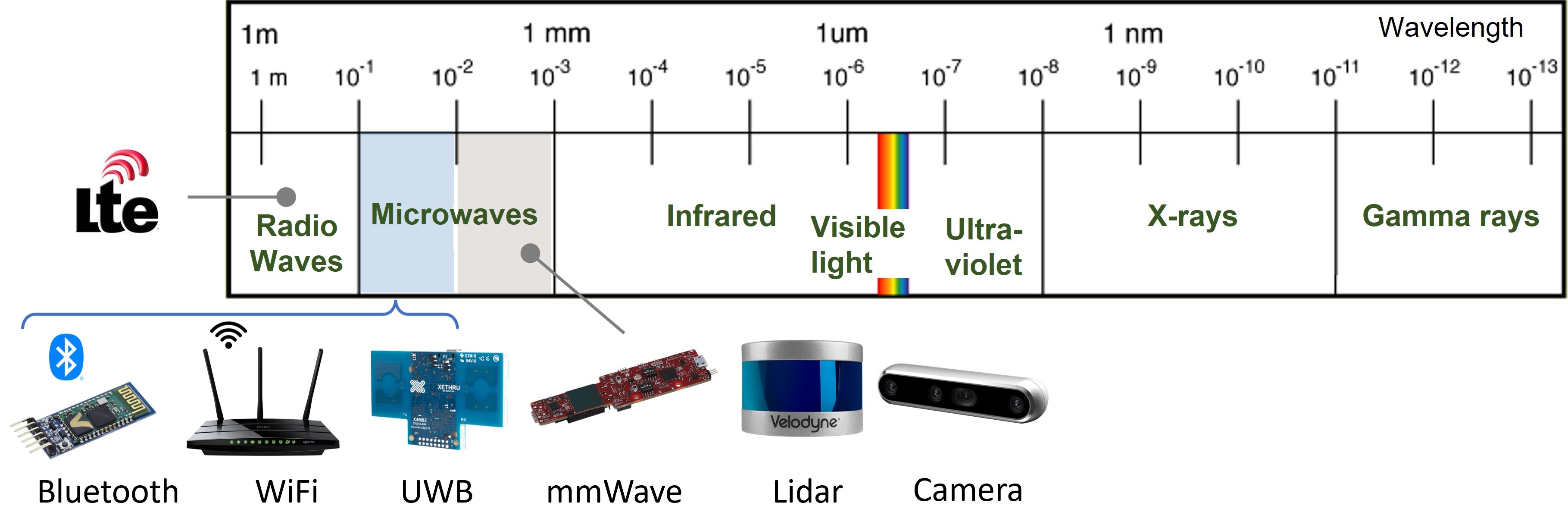}\label{envs_a}
	}
	
	\subfloat[][]{
		\includegraphics[height=0.92in]{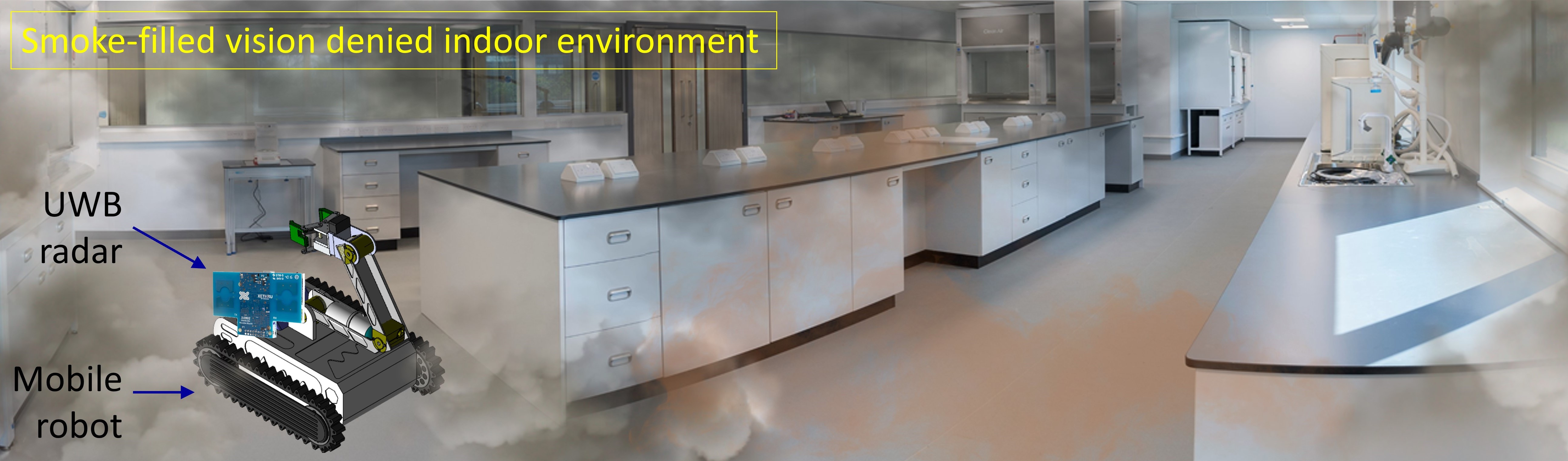}\label{envs_b}
	}
	
	\caption{(a) Popular sensors used in SLAM along the electromagnetic spectrum and the appropriate wavelength should be utilized based on the situation{/}environment. (b) A potential application of UWB radar SLAM: a rescue robot trying to carry out operations in a vision denied indoor environment.
	}
	\label{envs}
\end{figure}

At present, the most common exteroceptive sensors in SLAM are light detection and ranging (LiDAR) and optical camera systems\cite{RN70, RN3}. LiDAR uses laser beams to measure distance using time-of-flight, thus ambient lighting conditions do not affect its performance. LiDAR is widely used in state-of-the-art SLAM systems due to high accuracy and high sample density \cite{RN62, RN63}. Unlike LiDAR, typical visual cameras can be used in well-lit surroundings only. Furthermore, the light sources of both LiDAR and camera operate near the visible frequency region in the electromagnetic spectrum (see Fig. \ref{envs_a}). As a result, these optical measuring instruments are prone to fail under low visibility. This critically affects autonomous systems in adverse indoor conditions such as smoke and mirrors \cite{RN79, RN104, RN103}. 

Sound navigation and ranging (sonar) is a good alternative to perform under low visibility, especially in underwater environments \cite{RN75, RN107}. Sonar deploys acoustic waves to detect objects. However, both optical and sonar systems suffer from the refraction effect. This is a key issue while navigating in high temperature environments, especially when it comes to fire-rescue missions and operations inside mines. 

Considering these facts, recently, there have been several studies to exploit radio frequency (RF) bands instead of dedicated sensors for SLAM. \textcolor{black}{RF signals are not significantly affected by temperature or low visibility weather \cite{RN124}}. Hence, the robot can be localized by ubiquitous narrowband RF systems such as Wi-Fi and Bluetooth which use received signal strength indication (RSSI) with trilateration or comparing the radio fingerprints \cite{RN74, RN2}. However, these systems depend on the existing infrastructure such as Wi-Fi access points and Bluetooth beacons. On the other hand, wideband RF systems such as mmWave frequency-modulated continuous wave (FMCW) radar and pulse-modulated ultra-wideband (UWB) radar are also used in SLAM systems \cite{RN79, RN78, RN80, RN60, RN86}. 

This paper proposes \textcolor{black}{a} UWB radar-based system for a single mobile robot (mono-agent) to solve the SLAM problem in a vision denied indoor setting (see Fig. \ref{envs_b}) \textcolor{black}{while generating a landmark-based map and estimating the pose of the robot. Most importantly, the proposed method entirely depends on the onboard sensors of the robot. Furthermore, a simple outlier filtering scheme has been proposed before fusing the radar observations with odometry.}

The contributions of this paper are summarized as follows.
\begin{itemize}
	\item \textcolor{black}{We present an UWB radar-based anchorless SLAM system mobile robot which is capable of loop closure using point landmarks identified inside indoor environments;}
	\item We propose removal of detected false observations using density-based spatial clustering of applications with noise (DBSCAN) to detect point landmarks;
	\item \textcolor{black}{Experiments were performed to evaluate the proposed approach in different indoor environments;}
\end{itemize}

The rest of the paper is organised as follows. \textcolor{black}{Related work is discussed in Section \ref{RW}.} Section \ref{UWB} gives an insight to UWB including the radar sensor used in our approach. A detailed explanation of the proposed approach is given in section \ref{overview} which covers all the key aspects: signal acquisition and processing, \textcolor{black}{trilateration}, outlier removal using density-based clustering, and EKF SLAM with unknown correspondences. Section \ref{exp} validates the proposed approach in a real-world scenario. Finally, Section \ref{conc} concludes the paper.
\begin{figure}[t]
	\centering
	\subfloat[][]{
		\includegraphics[height=0.7in]{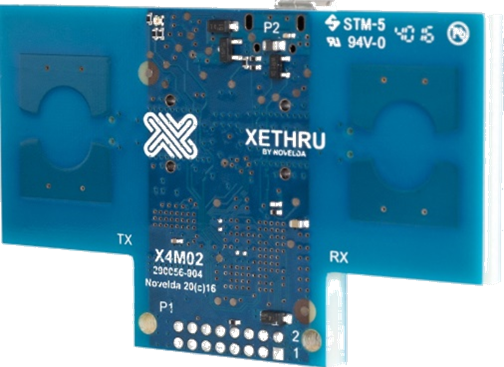}\label{sensor_a}
	}
	\subfloat[][]{
		\includegraphics[height=0.55in]{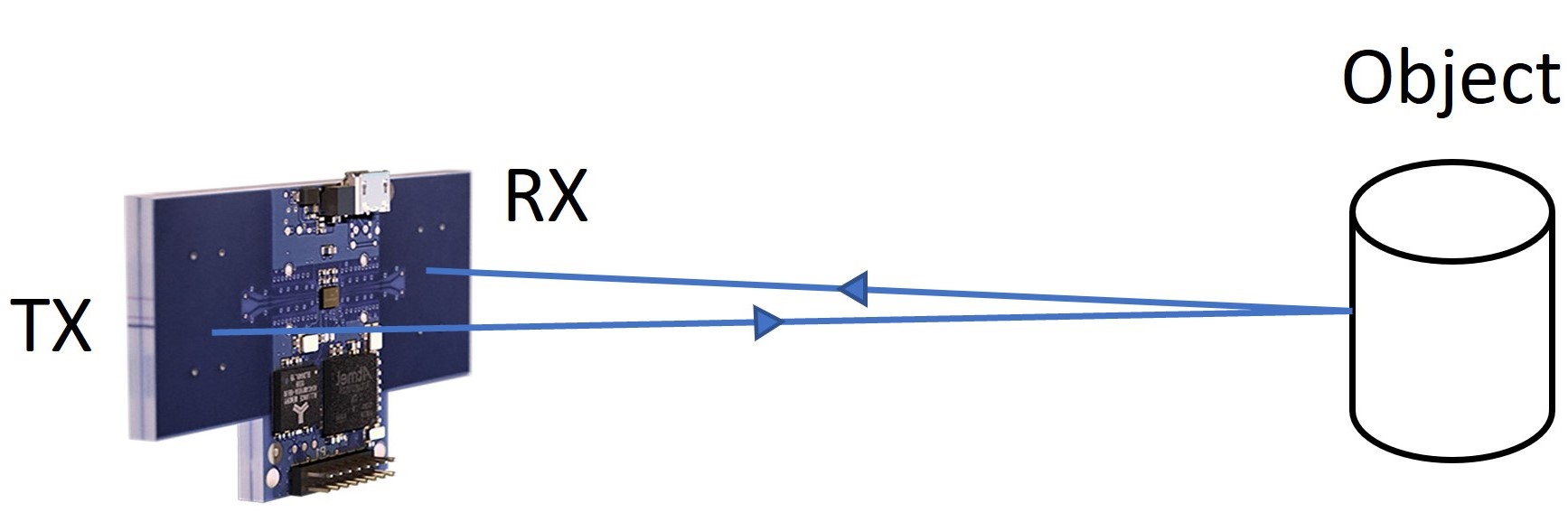}\label{sensor_b}
	}
	
	\caption{(a) X4M300 UWB radar sensor. \textcolor{black}{(b) X4M300 functions as a monostatic radar module.}
	}
	\label{sensor}
\end{figure}

\begin{figure}[t]
	\centering
	
	\subfloat[][]{
		\includegraphics[height=1.1in]{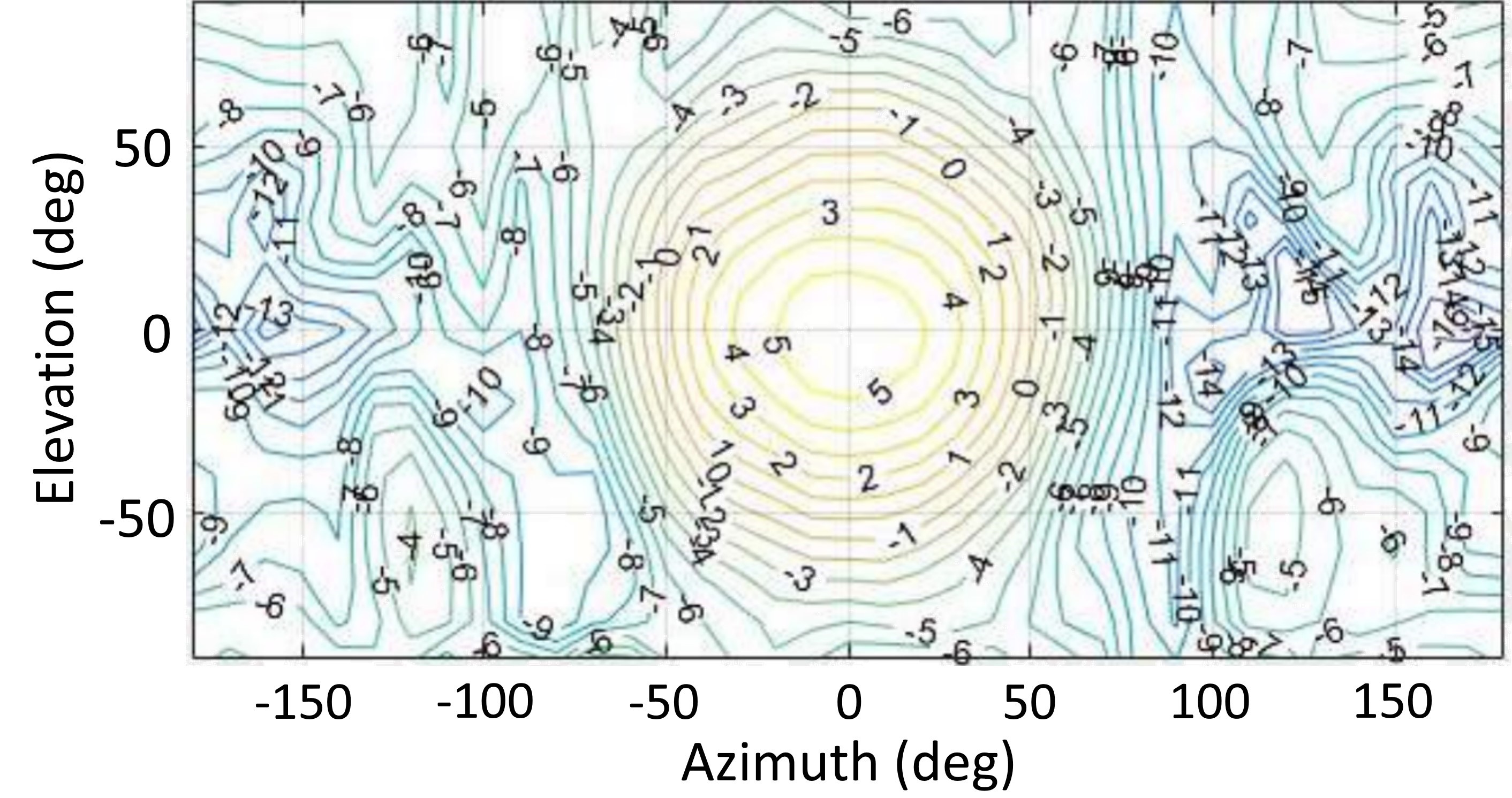}
	}
	\subfloat[][]{
		\includegraphics[height=1.1in]{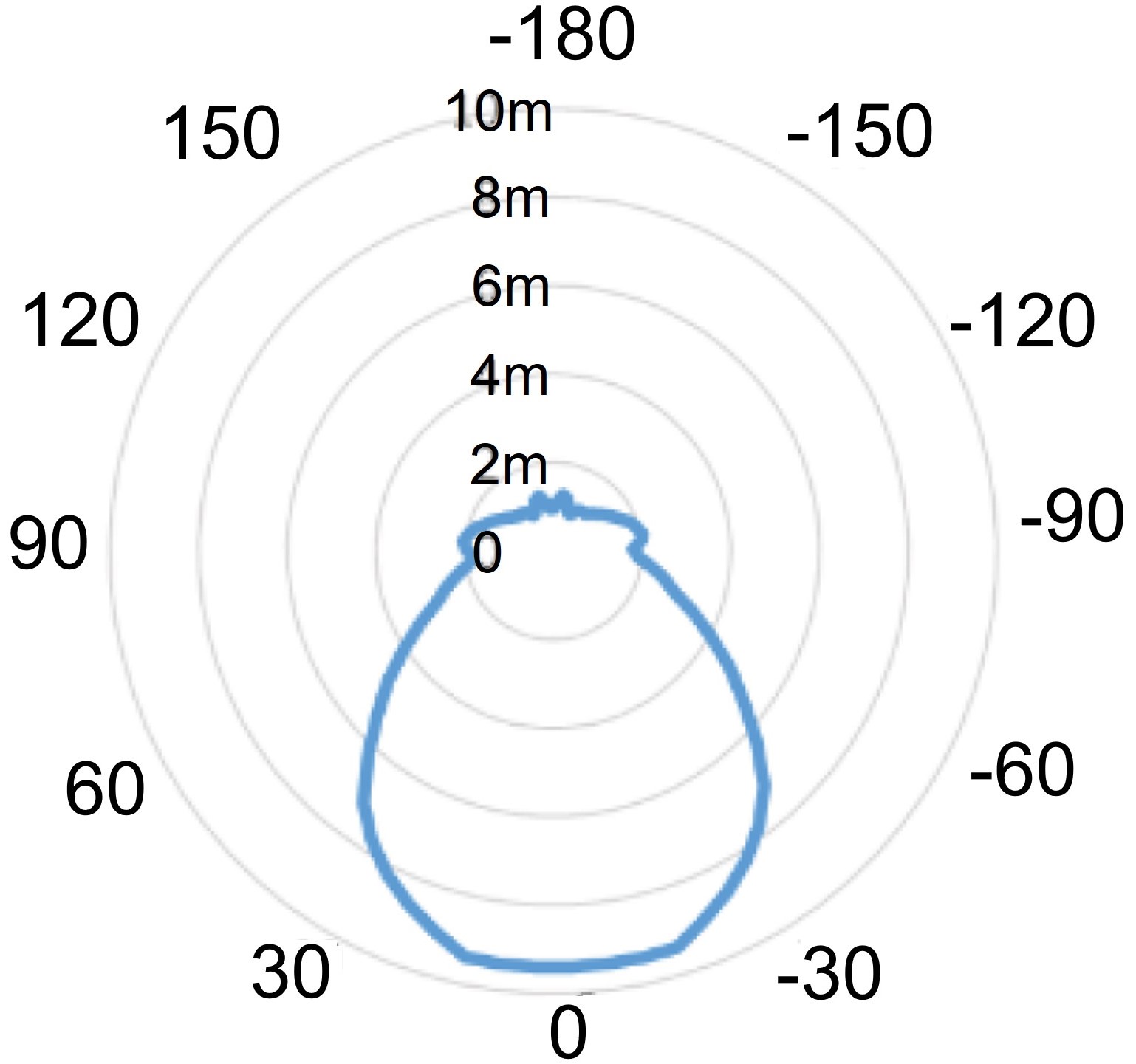}	
	}
	\caption{(a) X4M300 radiation pattern (max antenna gain: 5.915 dB) \textcolor{black}{(b) Azimuthal detection range of X4M300} \cite{x4m300, x4m02}.
	}
	\label{ant}
\end{figure}

\section{Related Work}\label{RW}

\textcolor{black}{
	Most UWB-based robot localization approaches use anchor-tag sensor configurations. In such systems,} a single or multiple anchors are usually fixed to known locations in the environment and the tags are mounted on the robots. The anchors are localizing the robot using two-way ranging (TWR)-based trilateration or by employing both TWR and angle-of-arrival information \cite{RN54, RN58, RN59}. The predominant limitation of anchor-tag approach is that the robot depends on the existing infrastructure (i.e. pre-installed anchors).

\textcolor{black}{
	The other major approach towards anchorless collaborative localization is by using peer-to-peer range measurements between multi-agents \cite{RN106, RN105}. In addition to localization, map building has also been studied under adverse conditions using UWB radar \cite{RN124}, \cite{RN128}. }

	There have been several studies which use a posteriori knowledge of the environment to perform UWB radar SLAM without anchors \cite{RN60, RN108}. These studies have been inspired from BatSLAM \cite{RN67} which is an adaptation of RatSLAM \cite{RN66}. The basic idea is to mimic the rat's brain where it gathers experiences about the surrounding by remembering its pose and distinct visual scenes. 
	Nonetheless, this method keeps a large dataset of previous observations (e.g. spectrograms \cite{RN60} or cochleograms \cite{RN67}) and requires high computational cost in order to compare the current place descriptor with the database. The major limitation is that the radar observations being susceptible to minor perturbations in the environment due to multipath propagation.
	
	Moreover, Deißler \textit{et al.} \cite{RN123, RN139} have used an onboard Bat-type UWB radar to perform SLAM using the detected features in the environment such as walls, points and corners. They have mainly focused on evaluating mapping accuracy using a precisely moving mechanical platform as well as odometry. However, they have not performed loop closure with their trajectory and the move-stop-rotation policy during the observations is quite tedious in practice. Moreover, their antenna setup takes a considerable footprint compared to off-the-shelf UWB radar modules.
	
	In contrast to the above methods, the proposed system uses \textcolor{black}{noisy} odometry data along with the observations from off-the-shelf UWB radar modules to realize anchorless SLAM. We have systematically evaluated both mapping and localizing performance of the proposed system inside indoor environments with loop closures. Leveraging on simplicity in design and versatility, UWB radar SLAM systems generate sufficiently accurate results, and is deemed to be low cost and consume less power compared to FMCW radar \cite{RN124}.
	
	 \begin{figure}[t]
		\centering
		{\includegraphics[scale=0.45]{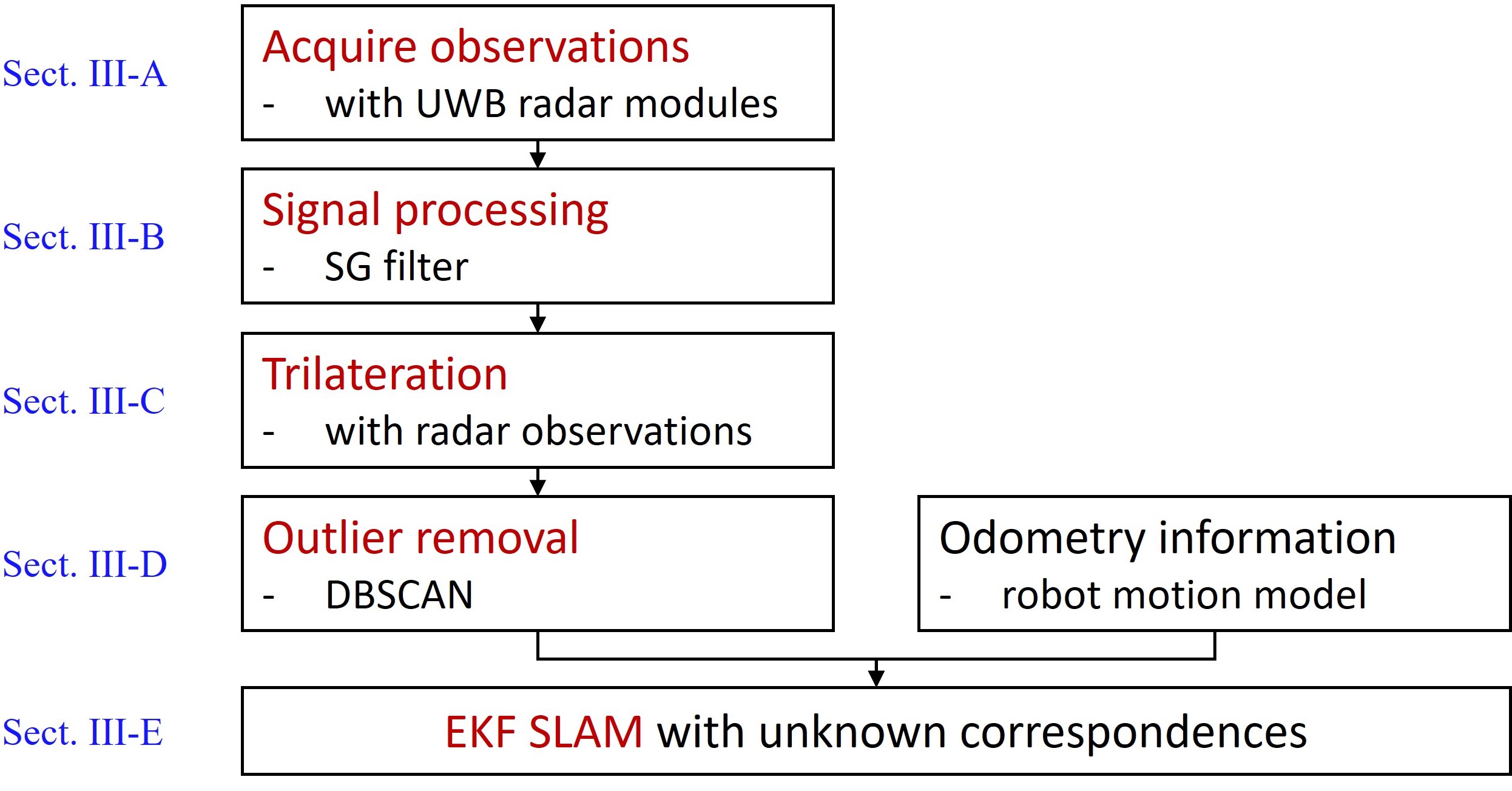}}
		\caption{Proposed UWB anchorless SLAM approach.}
		\label{ov}
	\end{figure}
	
	\begin{figure}[t]
		\centering
		\subfloat[][]{
			\includegraphics[scale=0.31]{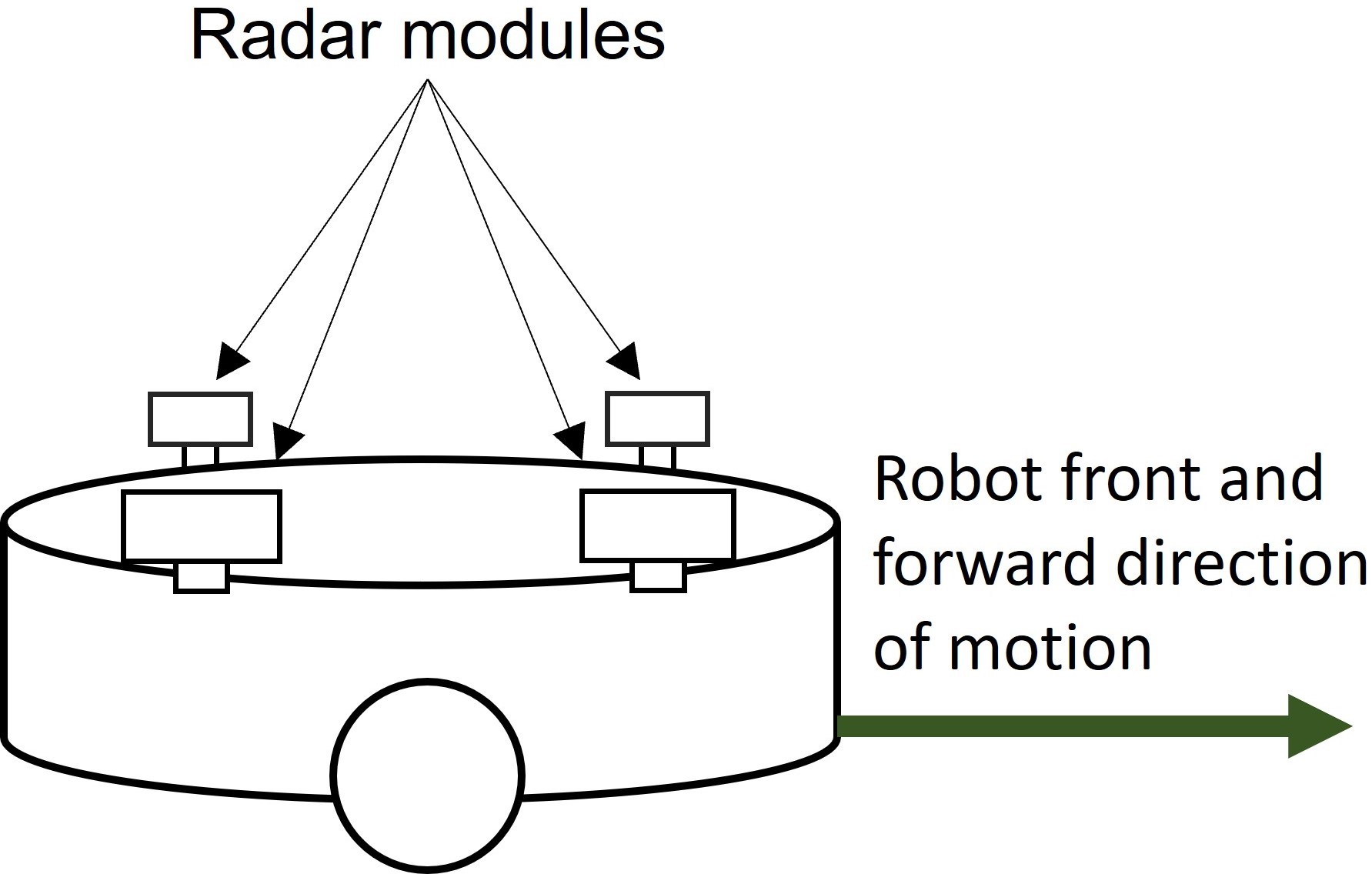}	
		}
		\subfloat[][]{
			\includegraphics[scale=0.31]{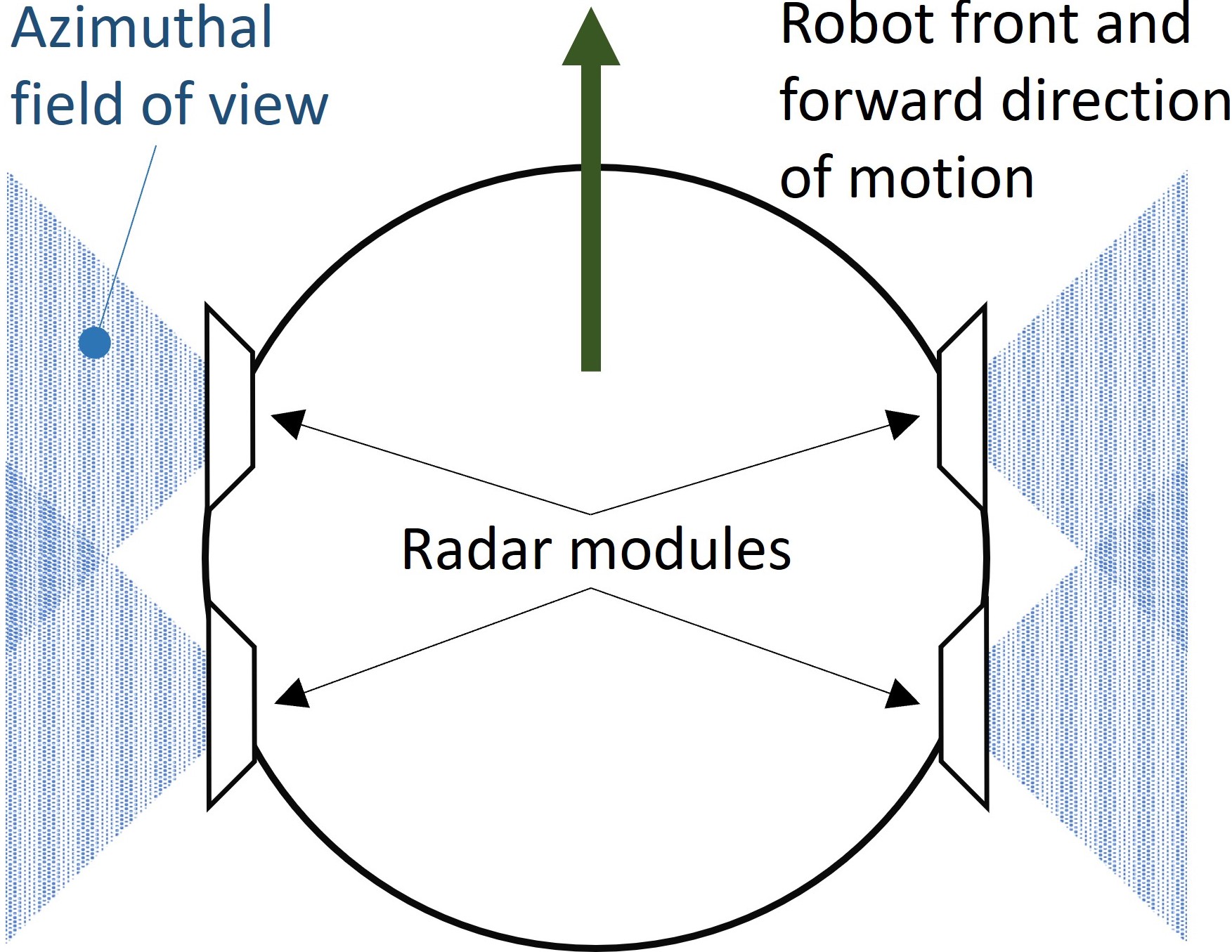}
		}
		
		\caption{Proposed radar sensor arrangement on a nonholonomic robot. (a) Side perspective view. (b) Plan view.
		}
		\label{arrangement}
	\end{figure}
	\begin{figure*}[t]
		\centering
		\subfloat[][]{
			\includegraphics[scale=0.67]{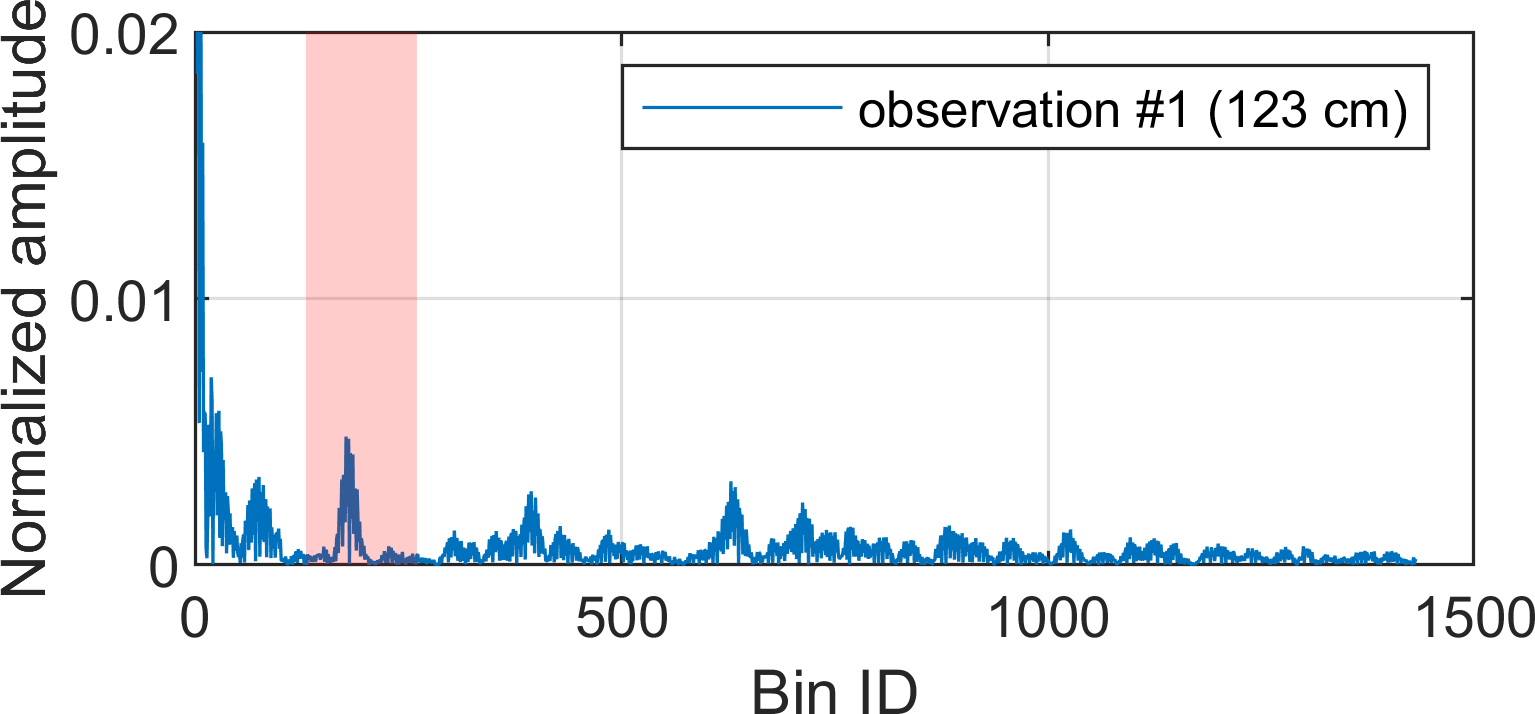}%
		}
		\hspace{0.5em}
		\subfloat[][]{
			\includegraphics[scale=0.67]{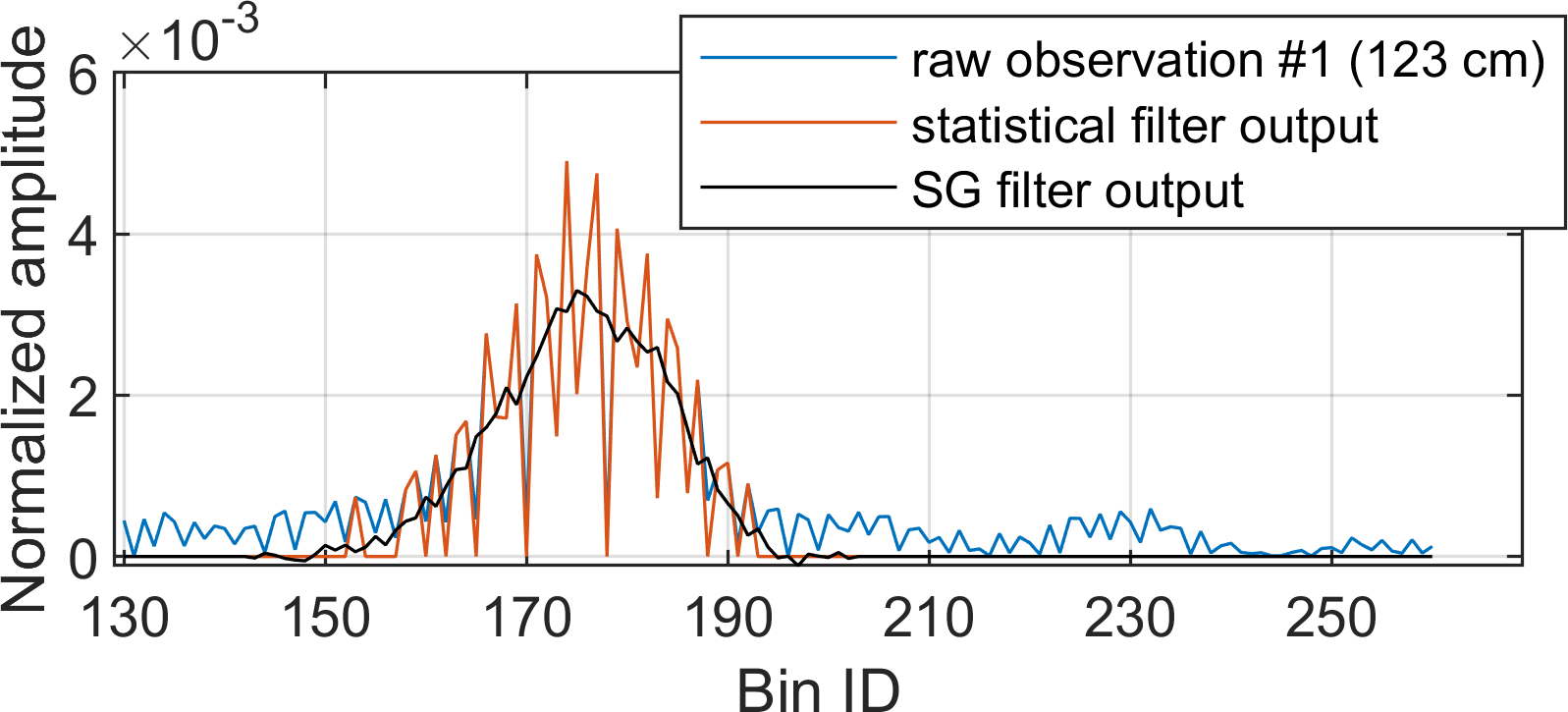}\label{signal_rf_b}
		}
		\hspace{0.5em}
		\subfloat[][]{
			\includegraphics[scale=0.67]{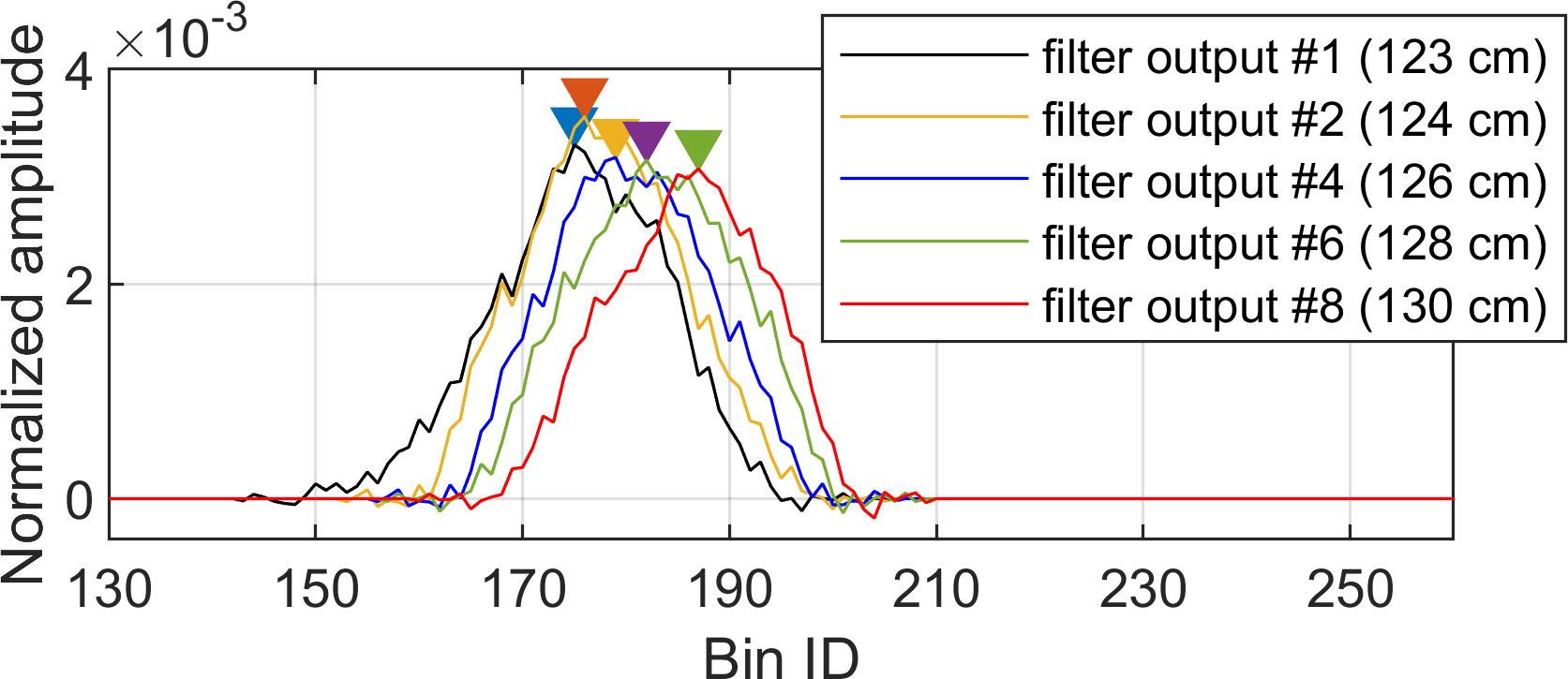}%
		}
		\caption{Raw radar observations from XeThru X4M300. \textcolor{black}{(a) Absolute values of the normalized aplitudes of UWB radar output. An object is somewhere in the shaded region. (b) Processed smooth signal using SG filter. (c) The object is moved away from the sensor and observations are taken at 1 cm intervals. As a result, the radar detections (i.e. identified peak amplitudes) are going away from the sensor.}}
		\label{signal_rf}
	\end{figure*}

\section{Background of Ultra-wideband}\label{UWB}

UWB is referred to in radio technology as a -10dB bandwidth that is larger or equal to 500 MHz or a fractional bandwidth greater than 20\%. This technology is generally used to communicate between two devices and for indoor positioning (i.e. TWR discussed earlier) \cite{RN64}.  However, this study mainly focuses on utilizing UWB radar towards anchorless mobile robot SLAM. UWB radar is deemed to have a high range resolution and ability to distinguish multiple targets due to high bandwidth and ultra-narrow pulses \cite{RN42}. Hence, UWB radar is also called impulse radar UWB (IR-UWB) and it is a prominent sensing principle for vital sign detection \cite{RN81, RN83, RN100}.

There are two types of radar: pulsed and continuous waves (CW) in which UWB falls under the pulsed category. Moreover, there are three types of radar systems based on antenna arrangement. i) Monostatic (i.e. transmitter and receiver collocated) ii) Bistatic (i.e. transmitter and the receiver are distant), and iii) Multistatic (i.e. a combination of monostatic and bistatic systems with a shared area of coverage) \cite{RN88}. 

The principle sensor of the proposed setup is XeThru X4M300 by Novelda (see Fig. \ref{sensor_a}). It is a monostatic IR-UWB radar module with inbuilt directional antennas for the transmitter and receiver (see Fig. \ref{sensor_b}). \textcolor{black}{Hence, this module is capable of ranging on its own\cite{RN124}, unlike tag-to-tag systems \cite{RN127}. As shown in the radiation patterns in Fig. \ref{ant}, the main lobe has an opening angle of 65$^\circ$ in both azimuth and elevation with a detection range up to 9.4 m and the back lobe effect is inconsiderable \cite{RN65, x4m300, x4m02}.}

\textcolor{black}{As mentioned in Section \ref{intro}, the other popular technology in the context of radar SLAM is mmWave (which belongs to FMCW).} However, UWB is deemed to have higher accuracy and signal to noise ratio (SNR) compared to FMCW. In addition, FMCW tends to attenuate more in indoor settings \cite{RN83, RN84}. Considering these facts, UWB is proposed in this study to achieve anchorless radar SLAM.

\section{System Overview}\label{overview}

This section explains all the building blocks of the proposed radar SLAM system. 
As shown in Fig. \ref{ov}, there are five major building blocks in the proposed SLAM system: acquire observations (radar signal acquisition) and preprocessing, then \textcolor{black}{trilaterating} the detections and remove outliers to obtain clear observations. Finally, those observations are combined with the odometry motion model in EKF SLAM.

\subsection{Radar Signal Acquisition}
In this paper, four XeThru UWB radar modules are attached to the either side of the robot as shown in Fig. \ref{arrangement}. The minimum requirement to proceed with \textcolor{black}{trilateration} is having at least two sensors at one side. Hence, additional sensors can be added if required to improve both accuracy and precision of the observations via \textcolor{black}{trilateration}. The detection zone of each module lies between 0.2 m and 9.4 m. Peak amplitudes in the radar readings correspond to radar observations from obstacles in its field of view \textcolor{black}{(FOV)} as shown in Fig. \ref{signal_rf}. The subsequent sections will explain the procedure and limitations encountered while acquiring the peak amplitudes. 

By default, the XeThru X4M300 has downconverted its range readings to baseband with 181 samples (bins) to ease post-processing.  \textcolor{black}{As a result, the spatial resolution of the baseband output is limited to 5.1 cm.}
In order to improve the resolution, the default settings were neglected, and the raw readings were taken into consideration (see Fig. \ref{signal_rf}). The raw readings have 1431 bins, so that the range is sampled at 6.4~mm intervals, which is much better than the baseband output. However, the raw signal is very noisy as expected. Thus, the next part explains the steps carried out to denoise the signal in order to obtain the necessary observations.
\setcounter{figure}{8} 
\begin{figure*}[b]
	\centering
	\subfloat[][]{
		\includegraphics[scale=0.5]{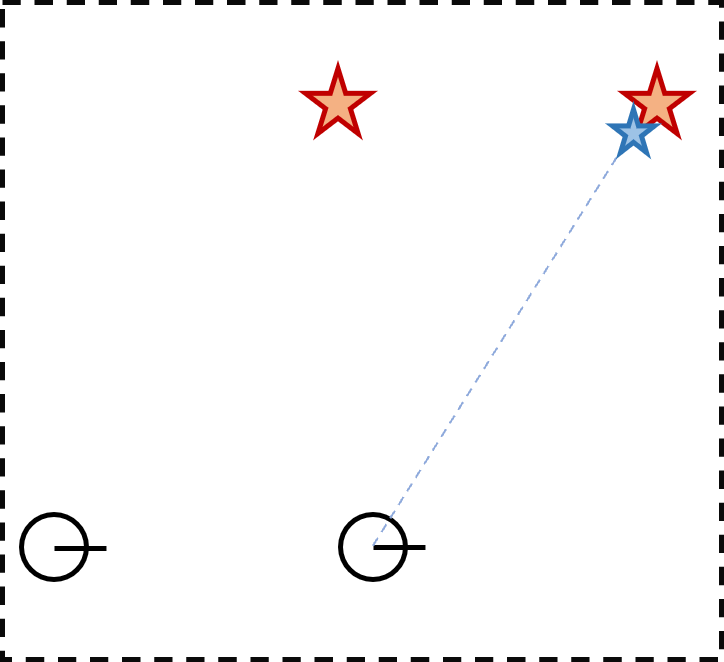}
	}
	\subfloat[][]{
		\includegraphics[scale=0.5]{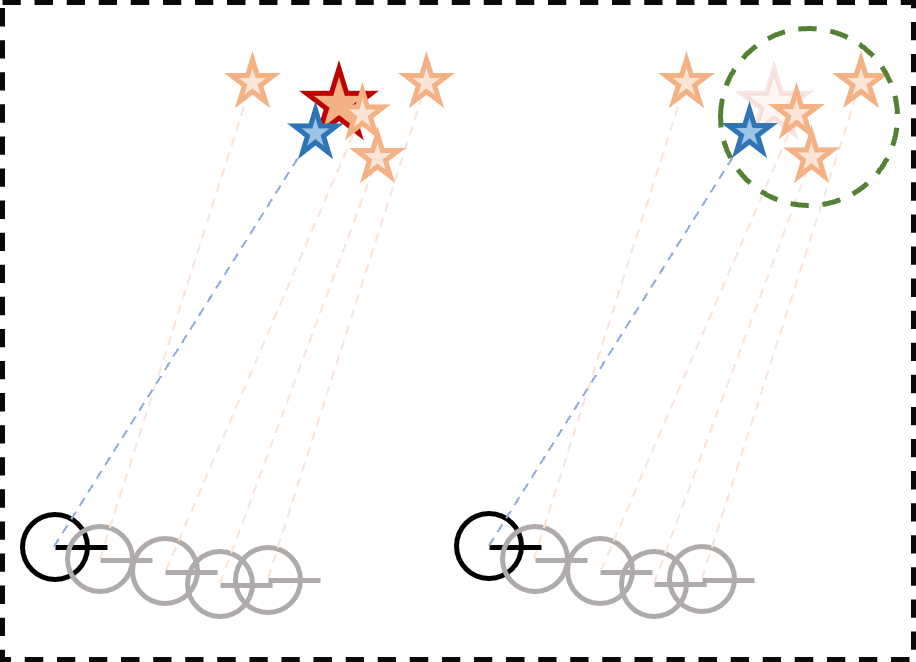}	
	}
	\subfloat[][]{
		\includegraphics[scale=0.5]{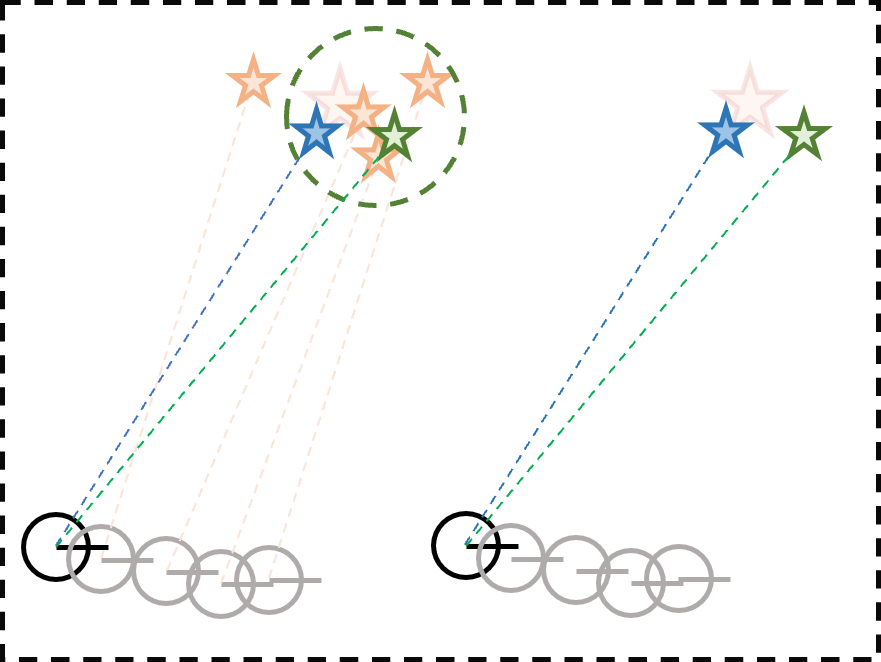}\label{dbs_c}
	}
	\subfloat[][]{
		\includegraphics[scale=0.5]{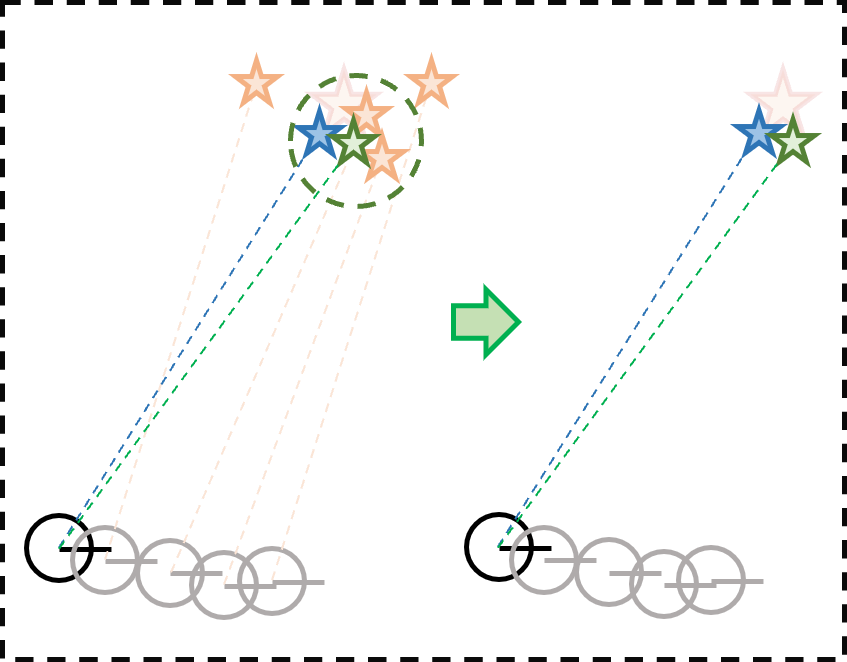}
	}
	\subfloat[][]{
		\includegraphics[scale=0.5]{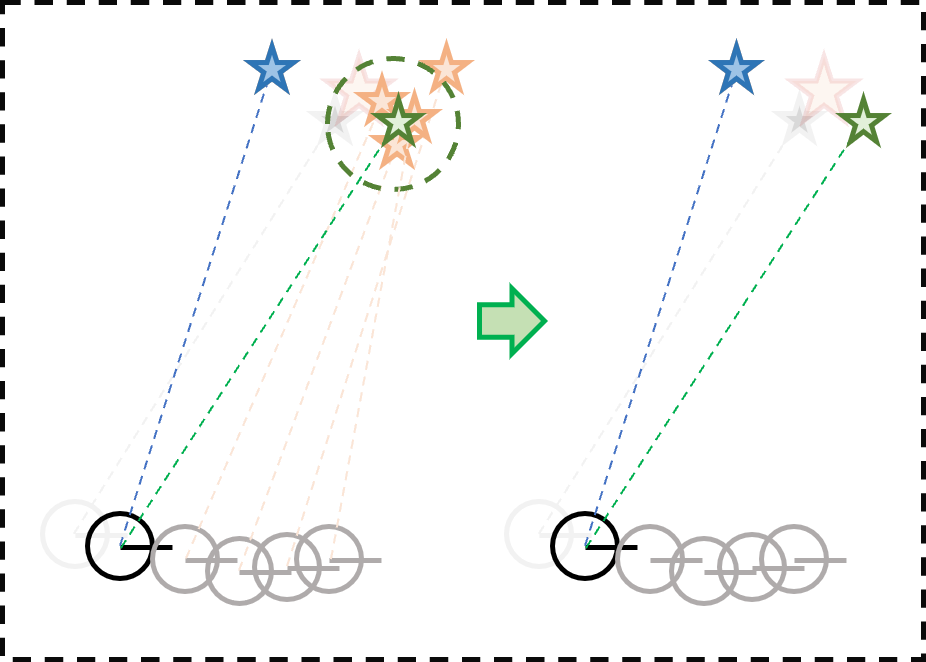}
	}
	\caption{Proposed semi-online DBSCAN-based outlier filtering algorithm. The robot pose is in the bottom, the real landmark position is the large star (black), and the observations are in small stars. (a) Robot observes a landmark in its \textcolor{black}{FOV} (current observation: blue star). (b) Basic intuition behind the proposed method: collecting a few observations (orange stars) ahead of the current observation and cluster all the observations using DBSCAN. (c) Assigned new observation (i.e. centroid of the clustered observations: green star) before and after clustering with a large DBSCAN search radius. (d) Observation before and after clustering with a small DBSCAN search radius. (e) The robot moves to the next EKF pose estimation and performs outlier removal again.
	}
	\label{dbs}  
\end{figure*}
\setcounter{figure}{6}
\begin{figure}[t]
	\centering
	\subfloat[][]{
		\includegraphics[scale=0.49]{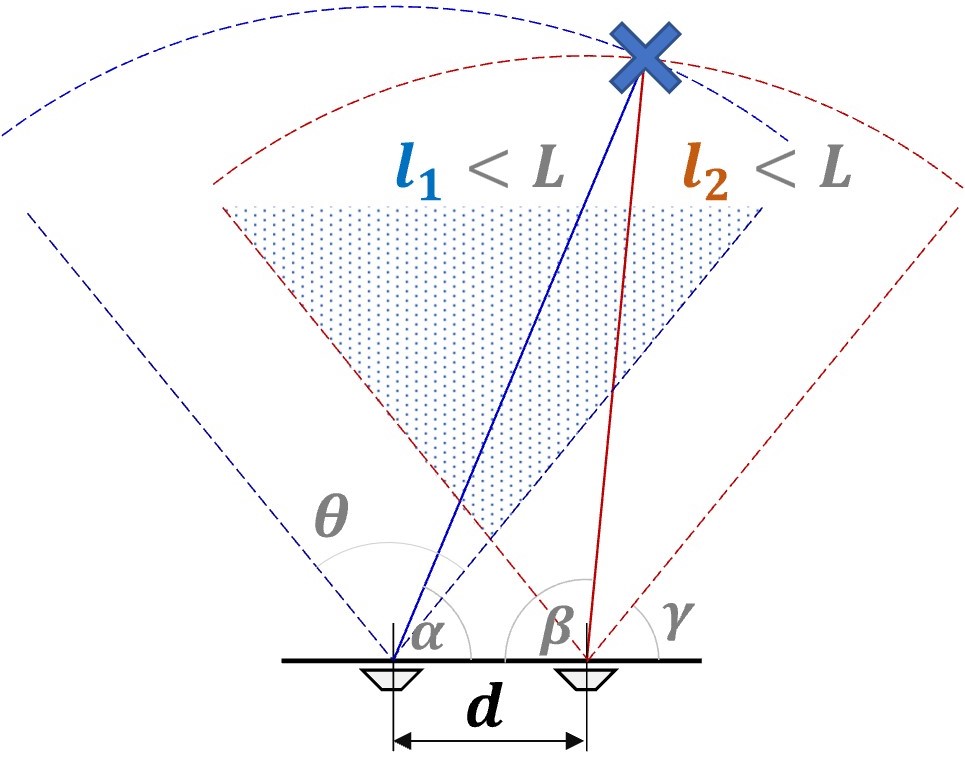}
	}
	\subfloat[][]{
		\includegraphics[scale=0.5]{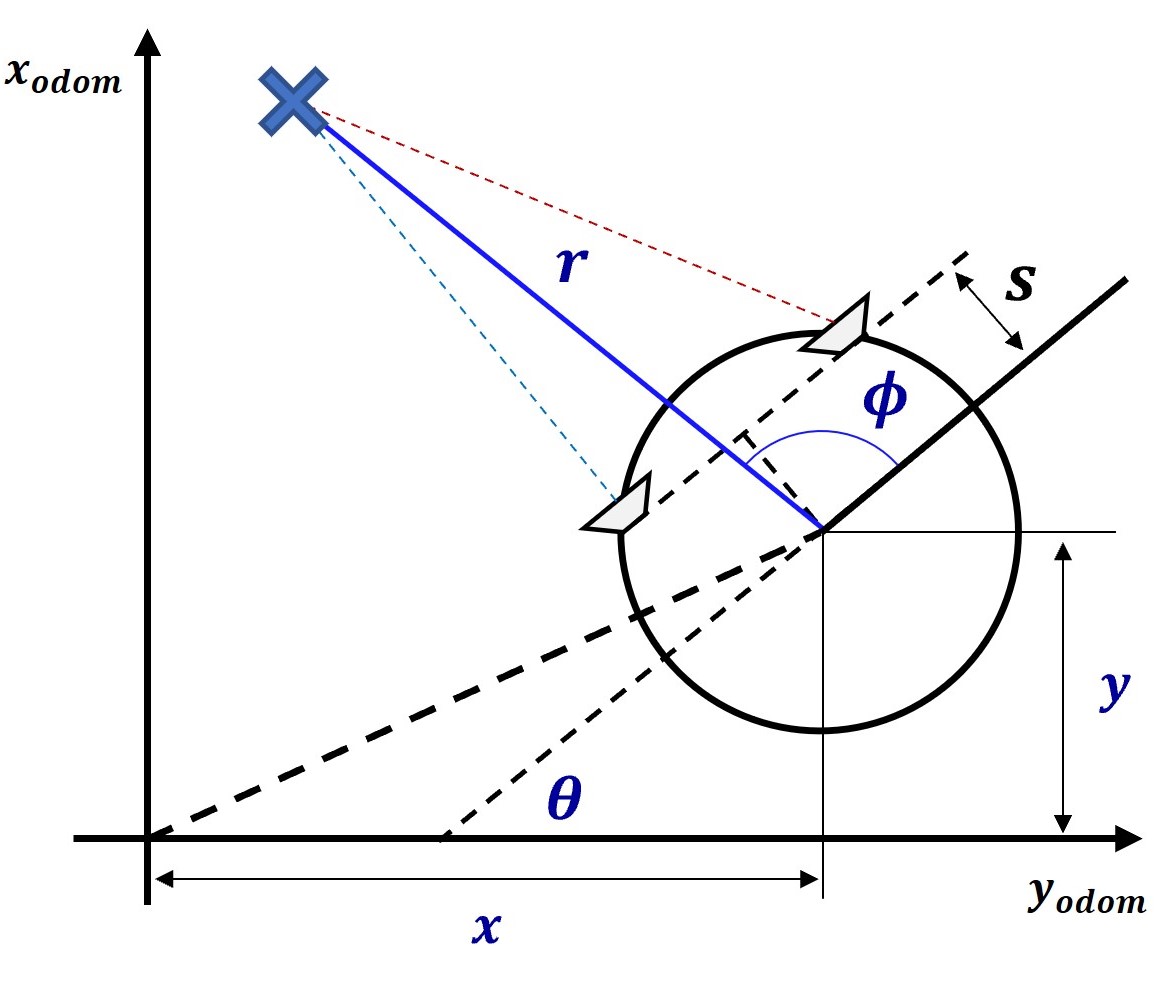}	
	}
	
	\caption{Determining the landmark location based on \textcolor{black}{trilateration}. (a) Geometric representation of the basic \textcolor{black}{trilateration} principle. (b) Schematic diagram of the critical geometric parameters with respect to the \textit{odom} frame. This figure depicts only the left-side UWB radar modules mounted on the robot.}
	\label{triang}
\end{figure}

\subsection{Signal Processing}
Two steps are proposed to smoothen the noisy signal and they are as follows:
\begin{enumerate}
	\item Basic statistical filter (all amplitudes under the mean amplitude will be removed. i.e. equalled to $0$)
	\item Then use Savitzky-Golay (SG) filter (fits a polynomial of order $n$ in a given frame length $m$)
\end{enumerate}

Initially, a basic statistical filter removes background white noise with near-zero amplitude. As a result, only the significant data points are sent to the SG filter, thus reducing the computational load. SG filter is a finite impulse response low-pass filter which fits $f\_len$ number of noisy data points to a polynomial of order $n$ using least squares method. SG filter can smoothen a noisy signal while preserving its shape and peaks \cite{RN101} and an example is illustrated in Fig. \ref{signal_rf_b}.

After smoothing the signal, local maxima (peak amplitudes) have to be found in order to obtain radar detections. There are two hyper-parameters to be considered: minimum peak height ($min\_ph$) and minimum peak prominence ($min\_pp$) (i.e. how much the peak stands out relative to other peaks). These parameters were adjusted using trial and error methods so that radar detections with low radar cross-sections (RCS) are omitted. As mentioned earlier, a peak corresponds to a target. Thus, two peaks from two nearby radar sensors are \textcolor{black}{trilaterated} to obtain 2D positional observations.

\subsection{\textcolor{black}{Trilateration}}\label{tri}
\textcolor{black}{Trilateration} is the fundamental principle behind most of the metrology systems in both 2D and 3D domains \cite{RN102}. The basic idea of \textcolor{black}{trilateration} is depicted in Fig. \ref{triang} using a pair of UWB radar modules. After detecting the peaks from each radar scan cycle, the corresponding ranges to the obstacles can be calculated (i.e. $l_1, l_2 < L$). Here, $L$ is a user defined constraint to omit observations beyond that range (i.e. similar to \textit{laser\_min\_dist} and \textit{laser\_max\_dist} in Hector SLAM). According to the Eqn. (\ref{gamma}), $\gamma$ depends on the azimuthal \textcolor{black}{FOV} ($\theta = 65^{\circ} $) of the sensor. $d$ is the distance between two radar sensor modules.
\begin{equation}\label{gamma}
	\textcolor{black}{\gamma = \frac{\left(\pi - \theta\right)}{2}}
\end{equation}
It is important that the intersection point (i.e. target) belongs to both radar sensors. Therefore, $\alpha$ and $\beta$ angles are compared with $\gamma$ and other intersections are omitted:
\begin{equation}
	\alpha = \frac{l_1^2 + d^2 -l_2^2}{\textcolor{black}{2l_1d}}>\gamma \text{ and } \beta = \frac{l_2^2 + d^2 -l_1^2}{\textcolor{black}{2l_2d}}>\gamma
\end{equation}
Radar readings to range $r$ and bearing $\phi$ conversion can be obtained via: 
\begin{equation}
	r = \sqrt{\left(l_1\cos\alpha - \nicefrac{d}{2}\right)^2 + \left(l_1\sin\alpha + s\right)^2}
\end{equation}
\begin{equation}
	\phi = \arctantwo{\left(l_1\sin\alpha + s,  l_1\cos\alpha - \nicefrac{d}{2}\right)}
\end{equation}
These $r$ and $\phi$ are observations (i.e. $z_t^i = [r_t^i$ $\phi_t^i]$) used in EKF SLAM together with the robot pose information $[x_t$  $y_t$  $\theta_t]^T$ from odometry data \cite{RN71}.

\subsection{Outlier Filtering using DBSCAN}\label{dbscan}

\begin{figure}[t]
	\centering
	\subfloat[][]{
		\includegraphics[height=1.485in]{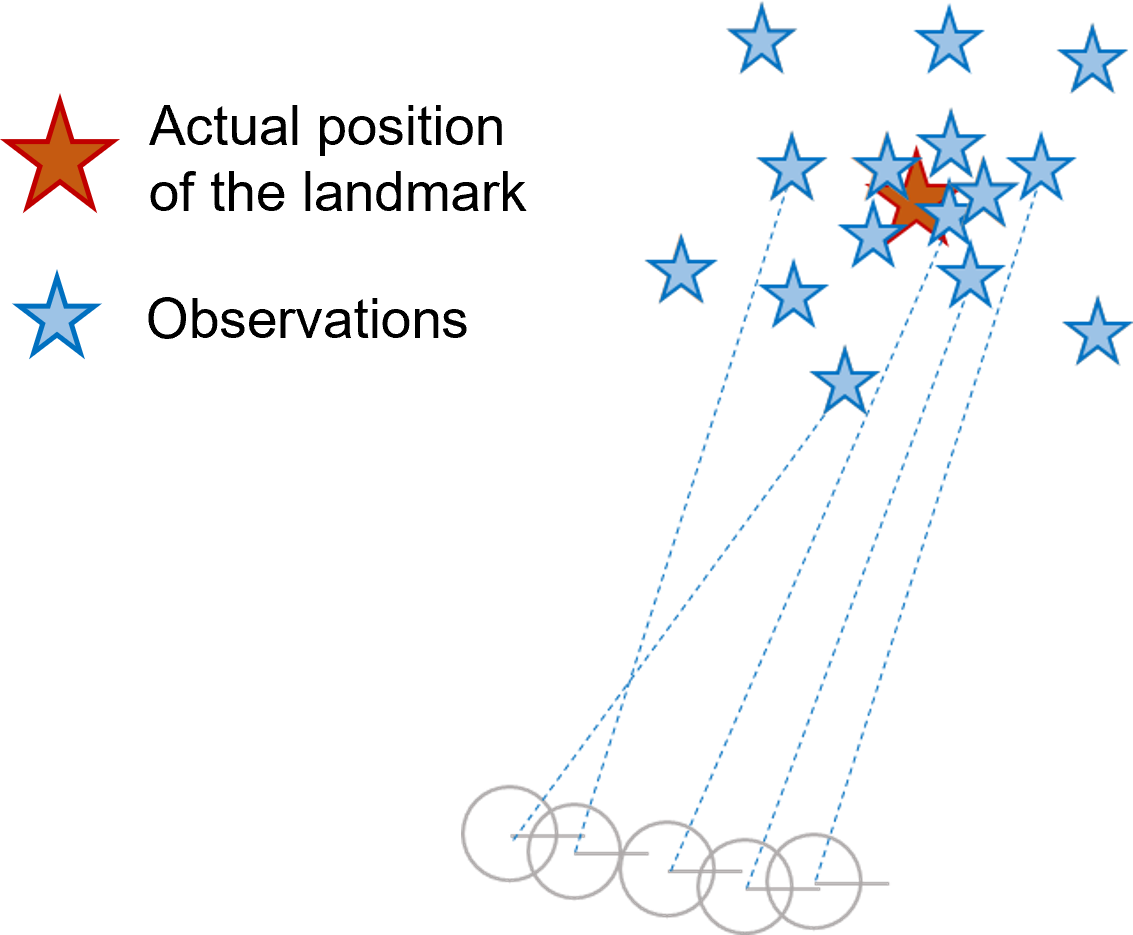}\label{outlier_a}
	}
	\subfloat[][]{
		\includegraphics[height=1.485in]{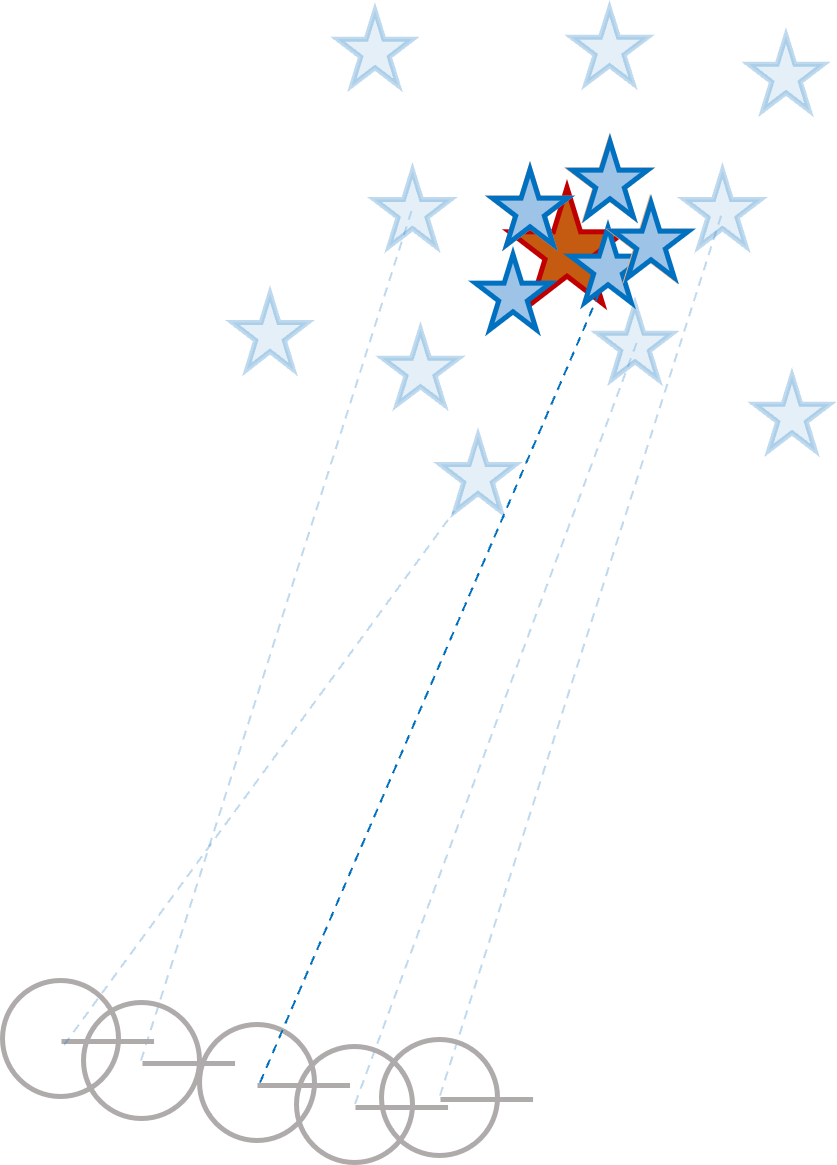}\label{outlier_b}
	}
	
	\caption{Illustration of the observations before and after outlier removal. The shaded stars are the omitted observations.} 
	\label{outlier}
\end{figure}

Although the preprocessing steps filter out the unnecessary observations in Section \ref{tri}, there still exist noisy sensory data which lead to outliers after \textcolor{black}{trilateration} as depicted in Fig.~\ref{outlier_a}. This figure shows known robot poses and corresponding observations. Obviously, the expected observations still have a certain uncertainty, but that will be handled by the EKF SLAM algorithm later.  In order to obtain a result similar to Fig. \ref{outlier_b}, this paper proposes density-based spatial clustering of applications with noise (DBSCAN) to filter the outliers as illustrated in Fig. \ref{dbs}. \textcolor{black}{DBSCAN has been used in FMCW radar to filter outliers in a point cloud \cite{RN196}. We are following a similar approach with UWB radar.}

The proposed DBSCAN-based method maintains a provisional observation list before they are sent to the EKF SLAM. The provisional observation list is clustered using DBSCAN and if a cluster is identified, its centroid is sent to EKF SLAM as the observation (see Fig. \ref{dbs_c}). This method has three hyper-parameters: size of the list of provisional observations (\textit{window\_size}), minimum number of points (i.e. observations) per cluster (\textit{min\_opc}), and the search radius (\textit{search\_rad}). 

\setcounter{figure}{9}
\begin{figure}[t]
	\centering
	\includegraphics[width=3.35in]{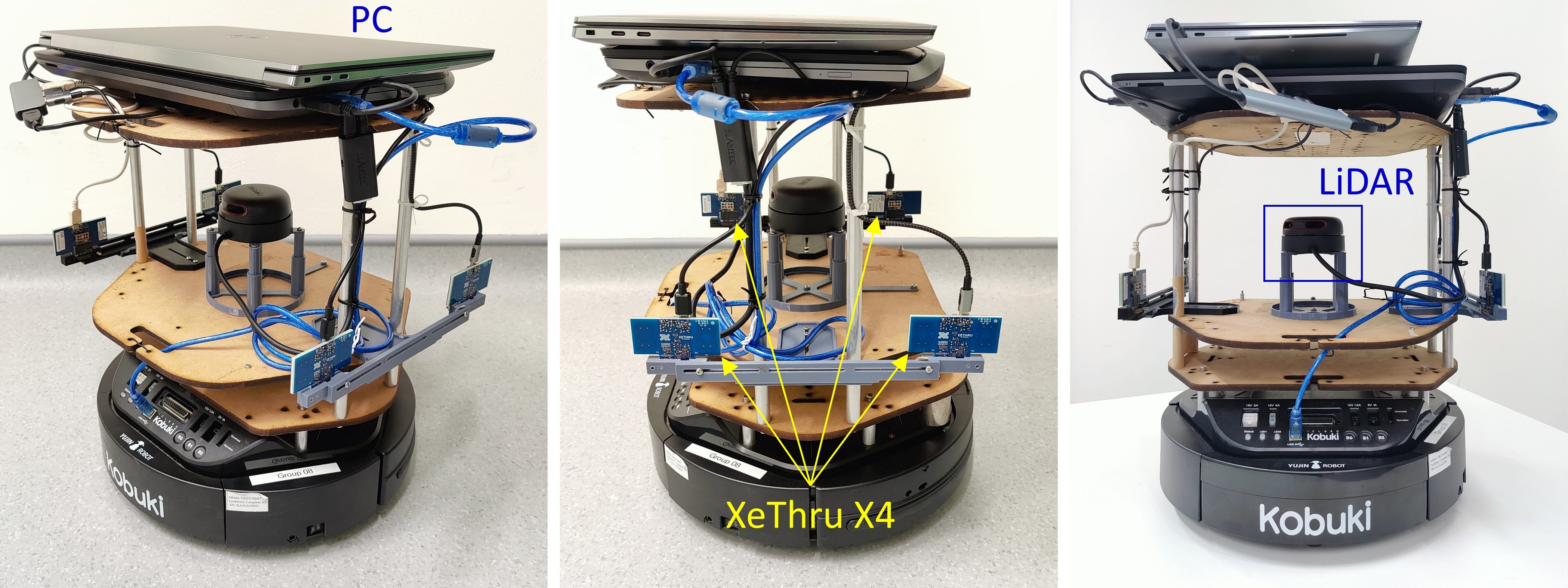}	
	\caption{ \textcolor{black}{Experimental setup with the TurtleBot2 research robot and the additional sensor modules. Note that the LiDAR is used to obtain the ground truth using Hector SLAM.}}
	\label{tbot}
\end{figure}
\begin{figure}[t]
	\centering
	\includegraphics[width=3.35in]{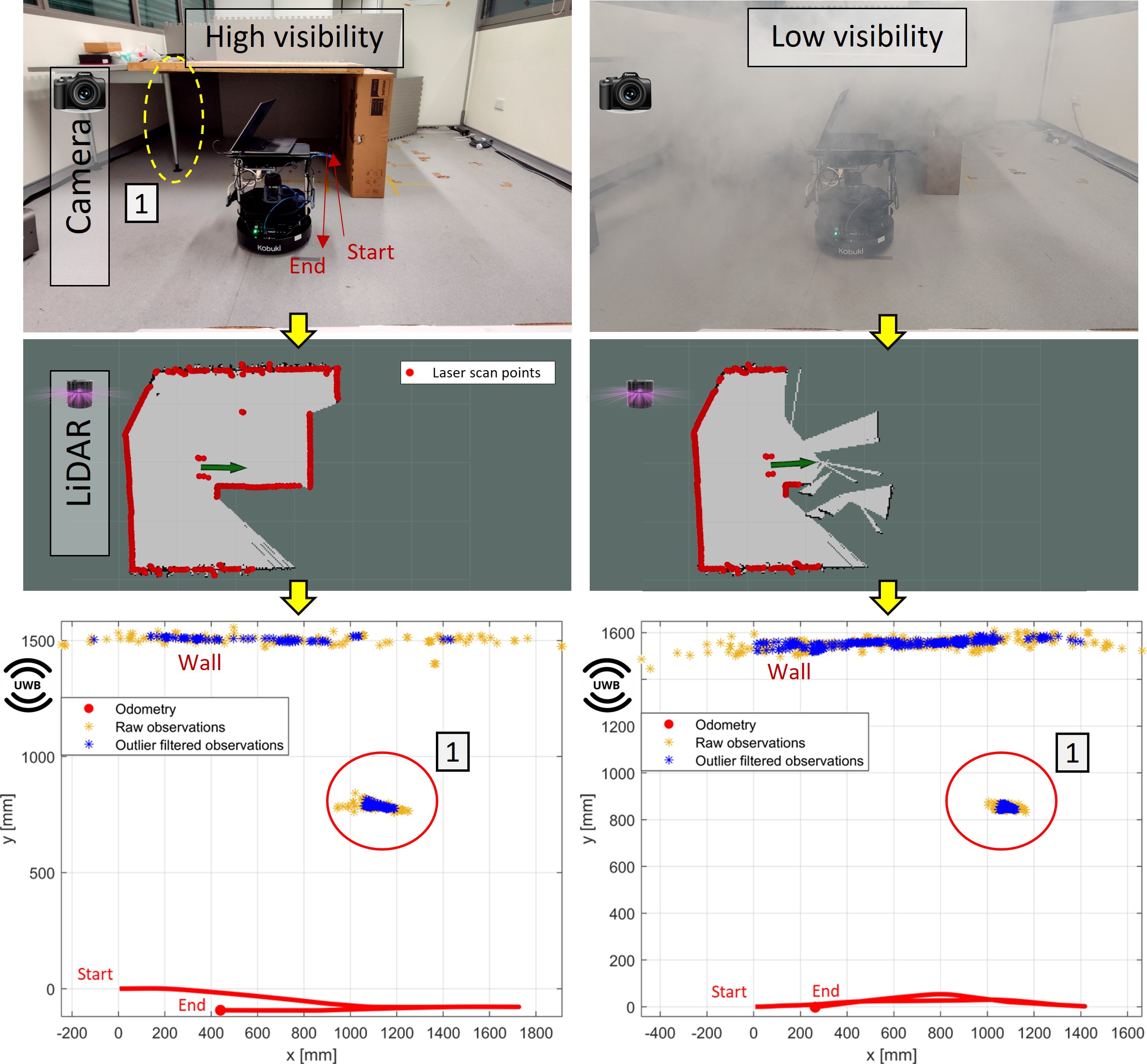}
	\caption{ \textcolor{black}{Both Camera and LiDAR fail under low visibility environment. However, UWB radar can observe the environment regardless of the visibility. Proposed DBSCAN-based outlier filtering method further refines the observations. }} 
\label{exp0}
\end{figure}

\textit{window\_size} is used to determine the number of states (with observations) of the robot to be collected before applying DBSCAN. A state is collected once the robot moves a threshold distance or an angle (\textit{min\_disp}) which is similar to \textit{map\_update\_distance\_thresh} and \textit{map\_update\_angle\_thresh} in Hector SLAM. The robot motion is determined using wheel odometry information. Hence, the higher the number of states to be collected (i.e. \textit{window\_size}), the more odometry-dependant the observations will be. On the other hand, the \textit{window\_size} cannot be decreased to a small number. If the \textit{window\_size} is reduced, there will be only a few observations, thus there will not be much difference between outliers and correct observations. 

\subsection{EKF SLAM}


\begin{algorithm}[t]
	\definecolor{black(pigment)}{rgb}{0, 0, 0}
	\color{black(pigment)}
	\caption{UWB radar SLAM}\label{startAlgo}
	\hspace*{\algorithmicindent} \textbf{Input:} motion uncertainty $R$, observation uncertainty $Q$, \textit{window\_size} for outlier removal $w$, observation update displacement threshold \textit{min\_disp}, EKF SLAM: initial mean $\mu$, initial covariance $\Sigma$, Mahalanobis distance threshold $\alpha$, Odometry data, Observations from UWB radar.\\
	\hspace*{\algorithmicindent} \textbf{Output:} robot pose and covariance.
	
	\begin{algorithmic}[1]
		\color{black(pigment)}
		\State $s \gets 1$\Comment{id of the estimated state}
		\State $i \gets 1$\Comment{id of the current state}
		\State $\text{Obs[]} \gets \text{empty list of observations}$
		\State $\text{Odom[]} \gets \text{empty list of odometry}$
		\State  $\textit{previous pose} \gets \text{read odometry}$\Comment{ $\text{Odom[0]} \gets \text{origin}$}
		
		\While{\textit{teleop}} \Comment{do while teleoperating the robot}
		\State $\textit{new pose} \gets \text{read odometry}$
		\State $\textit{disp} \gets \textit{new pose} - \textit{previous pose}$
		\If {$ \text{\textit{disp}} > \textit{min\_disp}$}
		\State $z_i \gets \text{observations from UWB radar modules}$
		\State Obs[$i$] $ \gets z_i$
		\State Odom[$i$] $ \gets \textit{new pose}$
		\State $i \gets i + 1$
		
		\If{$i > w$} \Comment{executes after collecting initial set of observations from $w$ number of poses}
		\State $\text{\textit{filt\_obs}} = DBSCAN(\text{Obs[$s : s + w - 1$]})$
		\Statex \Comment{filtered observations without outliers}
		\State \textit{u} $\gets$ odometry motion model
		\Statex	\Comment{calculated using  Odom[$s$] and Odom[$s - 1$]}
		\State $\bar{\mu}, \bar{\Sigma} \gets$ EKF prediction$(\mu, \Sigma, \textit{u}, \textit{R})$
		\State $\mu, \Sigma \gets$ EKF update$(\bar{\mu}, \bar{\Sigma}, \textit{Q}, \alpha, \text{\textit{filt\_obs}})$
		\State $s \gets s + 1$
		\EndIf
		\State $\textit{previous pose} \gets \textit{new pose}$
		\State\Return{robot pose $\gets \mu, \Sigma$}
		\EndIf
		\EndWhile
	\end{algorithmic}
	\label{algo1}
\end{algorithm}
\hspace{-1em}

EKF SLAM is used in this study in order to demonstrate the idea of UWB radar SLAM. EKF SLAM uses an extended Kalman filter (EKF) to estimate the pose of the robot and the positions of landmarks. 
When it comes to anchor-tag systems where anchors are landmarks, the anchors can embed unique identifications to inform the tag about the observation origin (known correspondences). However, as discussed in the previous sections, the final observation from the UWB radar does not contain any unique information about the landmark. Therefore, unlike in anchor-tag systems, the proposed method uses EKF SLAM with unknown correspondences algorithm. The Mahalanobis distance is calculated between each previously predicted landmark and a new observation before performing data association with respect to the threshold $\alpha$. \textcolor{black}{The steps to be followed while obtaining trilaterated radar observations are in Algorithm \ref{algo1}}.

\begin{figure}[t]
	\centering
	\subfloat[][]{
		\includegraphics[width=3in]{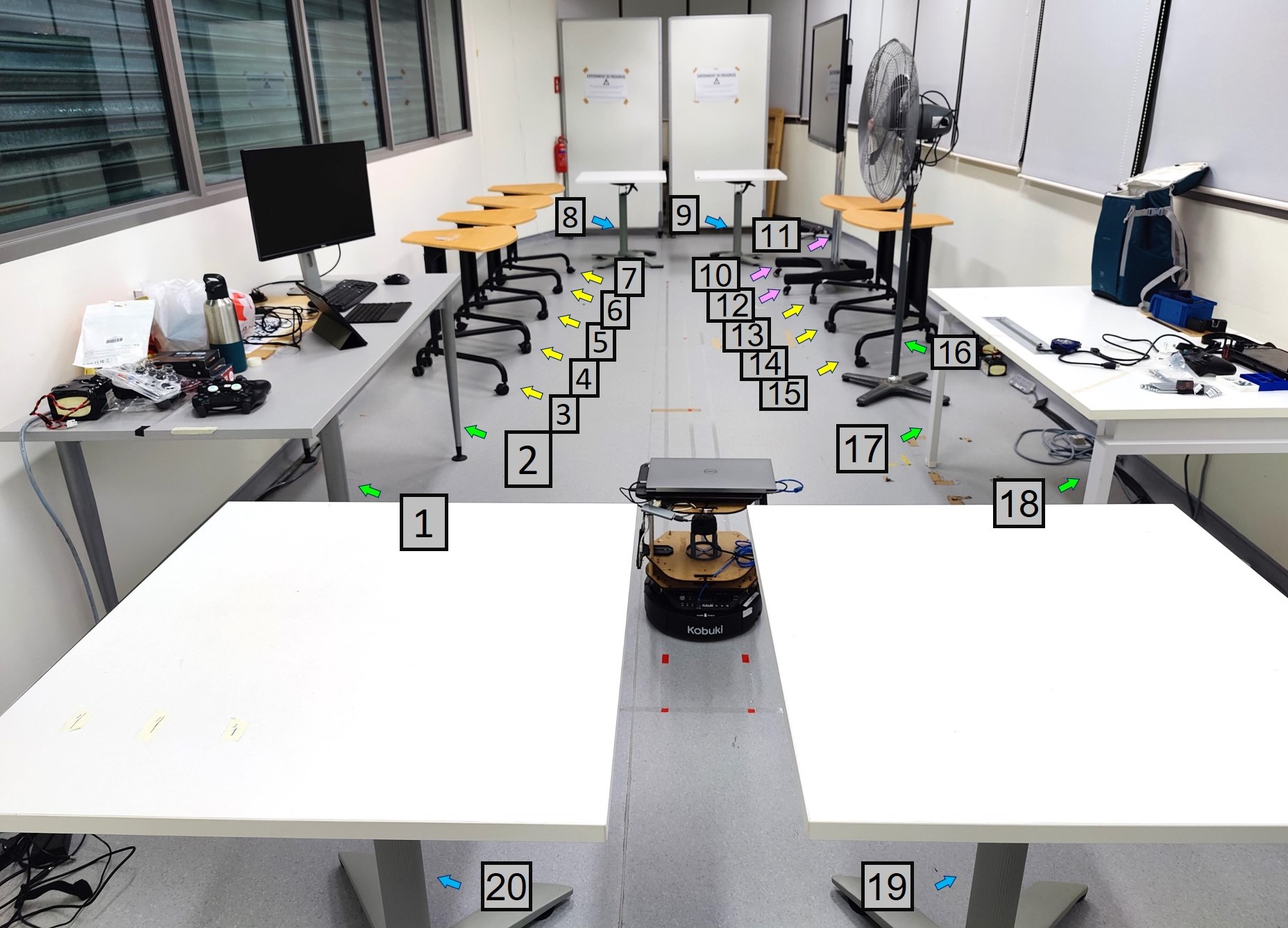}
	}
	
	\subfloat[][]{
		\includegraphics[width=3in]{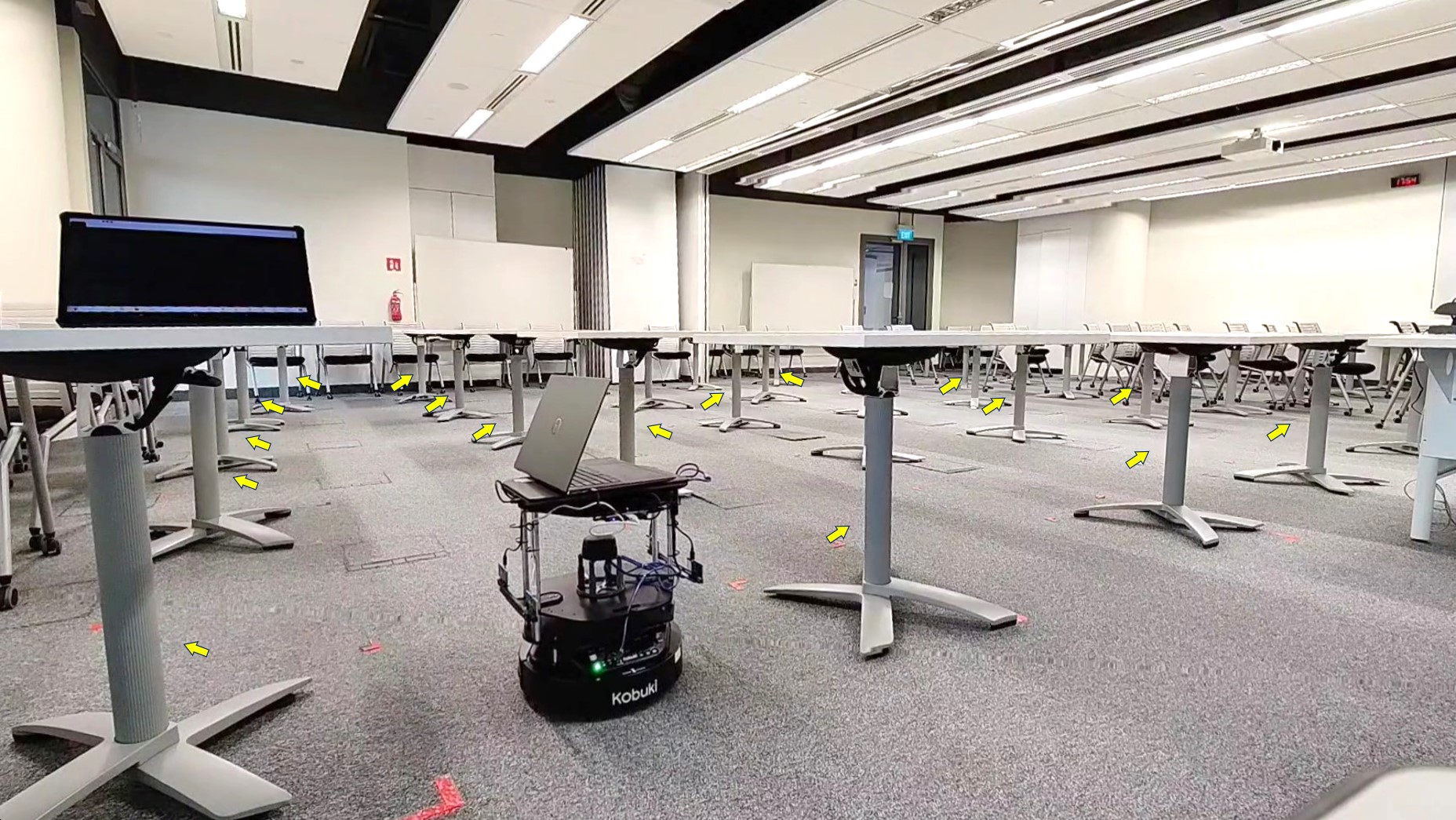}	
	}
	\caption{ \textcolor{black}{(a) Highly cluttered lab environment where the first experiment was conducted (b) Spacious classroom environment where the second experiment was conducted.}} 
\label{lab}
\end{figure}

\section{Experiments and Results}\label{exp}

A TurtleBot2 research robot is used in the experiment as shown in Fig. \ref{tbot}. RPLIDAR-A2 laser range scanner is mounted on top of the robot and is used to perform Hector SLAM in order to obtain the ground truth. The radar modules are operating at a frequency of 5 Hz while the robot is moving at 5 cm/s.

\textcolor{black}{\textit{1) Smoke Experiments:} In order to evaluate the performance of UWB radar in a vision-denied environment, we partially enclosed an indoor space with a point feature and performed Hector SLAM while obtaining UWB radar observations in clear visibility. After that, the space was filled with dense smoke and same steps were followed.
	As shown in Fig. \ref{exp0}, UWB radar has performed the same regardless of visibility, whereas LiDAR has performed poorly.} 

\textcolor{black}{
	Meanwhile, we examined the influence of the proposed filtering algorithm in Section \ref{dbscan} using wheel odometry as pose estimations.
	The filtered observations are more refined and clustered around the expected point landmarks (see Fig.~\ref{exp0}). During SLAM, the filter collects observations from \textit{window\_size} number of poses ahead of the current EKF pose estimation. These provisional observations are collected using only the odometry information. Hence, a small \textit{window\_size} is preferred to avoid odometry dependency on the filtering scheme. After collecting the provisional observations, there should be at least \textit{min\_opc} set of observations inside a radius of \textit{search\_rad}. \textcolor{black}{Hence, it is preferred to have high  \textit{min\_opc} and low \textit{search\_rad} to achieve maximum filter performance with minimum outliers. However, if these thresholds are initialized beyond practical extremities anticipating a high filter performance, the filter may end up omitting inliers as well.}
}

\textcolor{black}{
	The main objective in hyper-parameter tuning is to obtain low-noise  observations at the front-end. The detection range upper bound $L$ was chosen to detect only the nearby objects in order to avoid distant detections which may have resulted from multipath propagations, and to neglect the walls (i.e. line features). Later, we tuned \textit{min\_ph} and \textit{min\_pp} in order to identify prominent peaks from the UWB radar observations within the selected detection range, thus we could detect objects with large RCS.}

\begin{table}[t]
	\caption{Important parameters used in the proposed radar SLAM}\centering
	\begin{center}
		\setlength{\extrarowheight}{1pt}%
		\begin{tabular}{|p{0.075\textwidth}|p{0.085\textwidth}|p{0.25\textwidth}|}
			\hline
			\textbf{Parameter}&\textbf{Range}&\textbf{Remarks}\\
			\hline
			$\textit{f\_len}$& {$2m +1$\newline$m\in\mathbb{Z}^+$\newline$\textit{f\_len}=15$} & SG filter: The number of data points to\newline which the polynomial is fitted. Must be\newline an odd number. \\
			\hline
			$n$& \textit{n} < \textit{f\_len}\newline $n = 5$ & SG filter: Order of the polynomial to\newline be fitted. \\
			\hline
			$\textit{min\_ph}$&$5.5\times10^{-3}$ & Minimum peak height.\\
			\hline
			$\textit{min\_pp}$&$3\times10^{-3}$ & Minimum peak prominence.\\
			\hline
			$L$& $20 - 940$ cm \newline $L = 150$ cm& User-defined upper bound of the radar \newline range. Adjusted to detect only the\newline nearby targets.\\
			\hline
			$\theta$& $65^\circ$ & Azimuth opening angle.\\
			\hline
			$d$& $10 - 30$ cm \newline $d = 20$ cm & Distance between receiver antennas \newline (RX) of two radar modules. Depends\newline on the robot geometry. Higher is better.\\
			\hline
			$s$&$5 - 30$ cm \newline$s = 20$ cm& Perpendicular distance from the radars \newline to the base footprint. Depends on the\newline robot geometry.\\
			\hline
			$\textit{min\_disp}$& $5$ mm \textit{or}\newline 0.1 rad  & Updates the provisional observations\newline list and the SLAM once the robot\newline moves exceeding these predefined\newline minimum displacement thresholds.\\
			\hline
			$\textit{window\_size}$& $15$ & Size of the provisional observation list to which DBSCAN is applied. \\
			\hline
			$\textit{min\_opc}$&$9$ & Minimum observations per cluster \textcolor{black}{for \newline DBSCAN-based outlier filtering}.\\
			\hline
			$\textit{search\_rad}$& $20$ cm& Search radius of clusters with \textit{min\_opc}.\\
			\hline
			$R$& \text{ }
			\newline 
			\resizebox{0.09\textwidth}{!}
			{$
				\begin{bmatrix}
					\sigma_x^2 & 0 & 0 \\
					0 & \sigma_y^2 & 0\\
					0 & 0 & \sigma_\theta^2\\
				\end{bmatrix}
				$}
			
			\textcolor{black}{
				\resizebox{0.95\hsize}{!}
				{$ $\newline
					\text{ }\newline}
				\resizebox{0.95\hsize}{!}
				{$\newline\sigma_\theta \text{ = } 0.02 \text{ rad\newline }$}}
			& Motion uncertainty of the robot between two poses. Depends on \textit{min\_dist}. Larger the \textit{min\_dist} is, larger the uncertainty will be, and vice versa. \newline
			\textcolor{black}{Lab experiment: $\sigma_x = \sigma_y = 0.1$ mm \newline
				Classroom exp: $\sigma_x = \sigma_y = 1$ mm}  \\
			\hline
			$Q$& \text{ }
			\newline 
			\resizebox{0.06\textwidth}{!}
			{$
				\begin{bmatrix}
					\sigma_r^2 & 0\\
					0 & \sigma_\phi^2\\
				\end{bmatrix}
				$}
			& Observation uncertainty. Large uncertainty due to inherent measurement\newline noise. \textcolor{black}{{$\sigma_r \text{ = } 15\text{ cm, } \sigma_\phi \text{ = } 1 \text{ rad\newline }$}}\\
			\hline
			$\alpha$& \textcolor{black}{$0.5 - 3$} & Mahalanobis distance threshold.
			\textcolor{black}{\newline 
				Lab experiment: $\alpha_1 = 0.6$ \newline
				Classroom exp: $\alpha_2 = 1.6$} \\
			\hline
		\end{tabular}
		\label{tab1}
	\end{center}
\end{table}

\setcounter{figure}{12}
\begin{figure*}[t]
	\centering
	\subfloat[][]{
		\includegraphics[scale = 0.42]{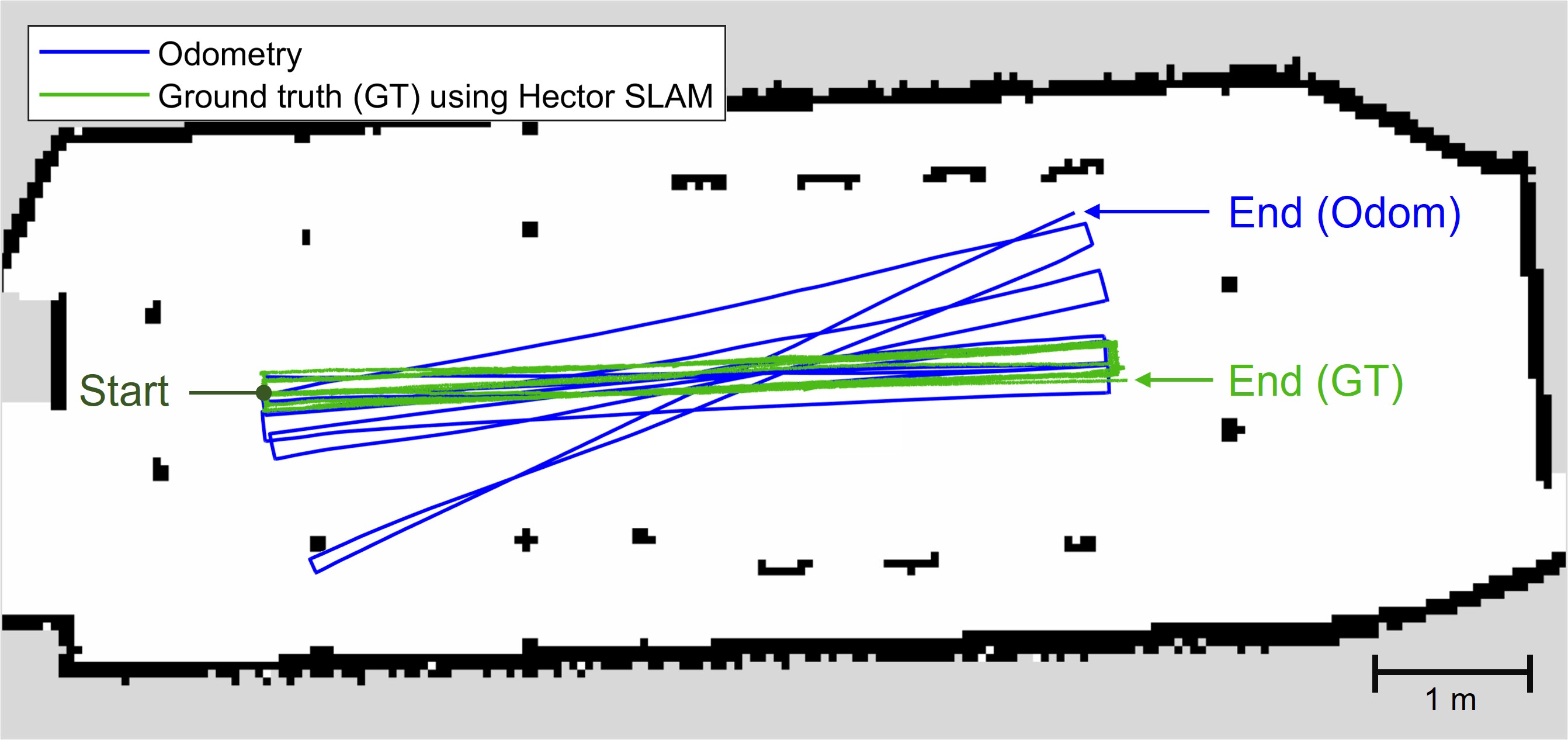}
		\label{exp1a}
	}
	\hspace{0.1em}
	\subfloat[][]{
		\includegraphics[scale = 0.42]{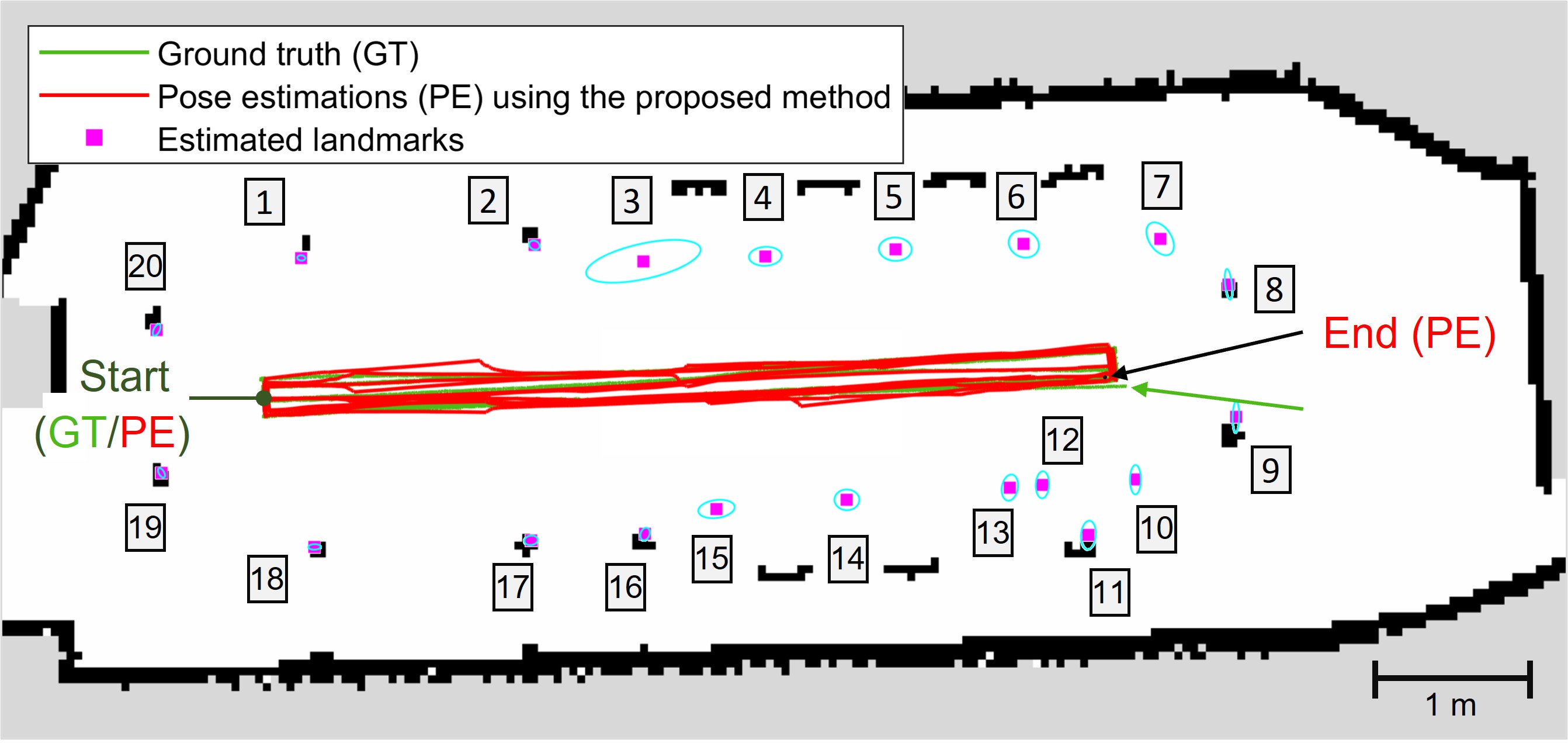}
		\label{exp1b}
	}
	\caption{ \textcolor{black}{Experiment 1 in the lab environment (a) The robot was teleoperated in a square-shaped trajectory multiple times in both clockwise and anticlockwise directions. The wheel odometry has a considerable drift compared to the ground truth. (b) The estimated robot poses and the map constructed by the proposed UWB radar SLAM method. The error ellipses around the estimated landmarks illustrate uncertainties.}} 
\label{exp1}
\end{figure*}

\textcolor{black}{
Table \ref{tab1} summarises the hyper-parameters used during the experiments included in this paper. 
It is important to notice that all these parameters are interrelated. If the proposed system is to be recreated using different UWB radar modules or in an environment consisting of objects with low RCS, all these parameters may have to be retuned. However, the intuition behind parameter selection is essential to improve the overall performance of the proposed SLAM system.}

\textcolor{black}{\textit{2) SLAM Experiments: }}\textcolor{black}{Two environments were selected to evaluate the proposed UWB radar SLAM: a highly cluttered lab environment and a classroom as shown in Fig. \ref{lab}.} These environments consist of point features with large RCS, such as vertical metal rods.   
Radar SLAM creates a landmark-based map while estimating the robot pose, whereas Hector SLAM creates a 2D occupancy grid map.  \textcolor{black}{The \textit{slam\_out\_pose} published by Hector SLAM was taken as the ground truth reference.} \textcolor{black}{We evaluate the accuracy of the proposed system with root mean squared (RMS) absolute trajectory error (ATE) after aligning the estimated poses with ground truth using the iterative closest point (ICP) algorithm.}

In the first experiment, the robot was teleoperated in a square-shaped trajectory inside a lab. The ground truth and wheel odometry are superpositioned in Fig. \ref{exp1a}.
Compared to the ground truth, the odometry has shifted anticlockwise during the experiment. Although the UWB radar SLAM initially tends to follow the odometry, it finally identifies previously known landmarks and corrects its pose. Both ground truth and radar SLAM have started and stopped at almost the same position (see Fig. \ref{exp1b}). Compared to the ground truth, the estimated positions had a ATE of 12~mm.

Notice that the landmarks: \#3-7, \#10 and \#12-15 were not captured by Hector SLAM, but they were detected by radar SLAM (see Fig. \ref{exp1b}). However, rest of the landmarks coincide with their ground truth with a very small uncertainty. The reason is that the RPLIDAR-A2 captures observations on a 2D plane whereas the UWB radar antenna has a cone-shaped main lobe in 3D space. As a result, the UWB radar has been able to detect observations which were not visible to the 2D LiDAR. This is one of the advantages of radar, despite unnecessary targets can also be detected in its \textcolor{black}{FOV}.

\begin{figure}[t]
	\centering
	{\includegraphics[scale = 0.41]{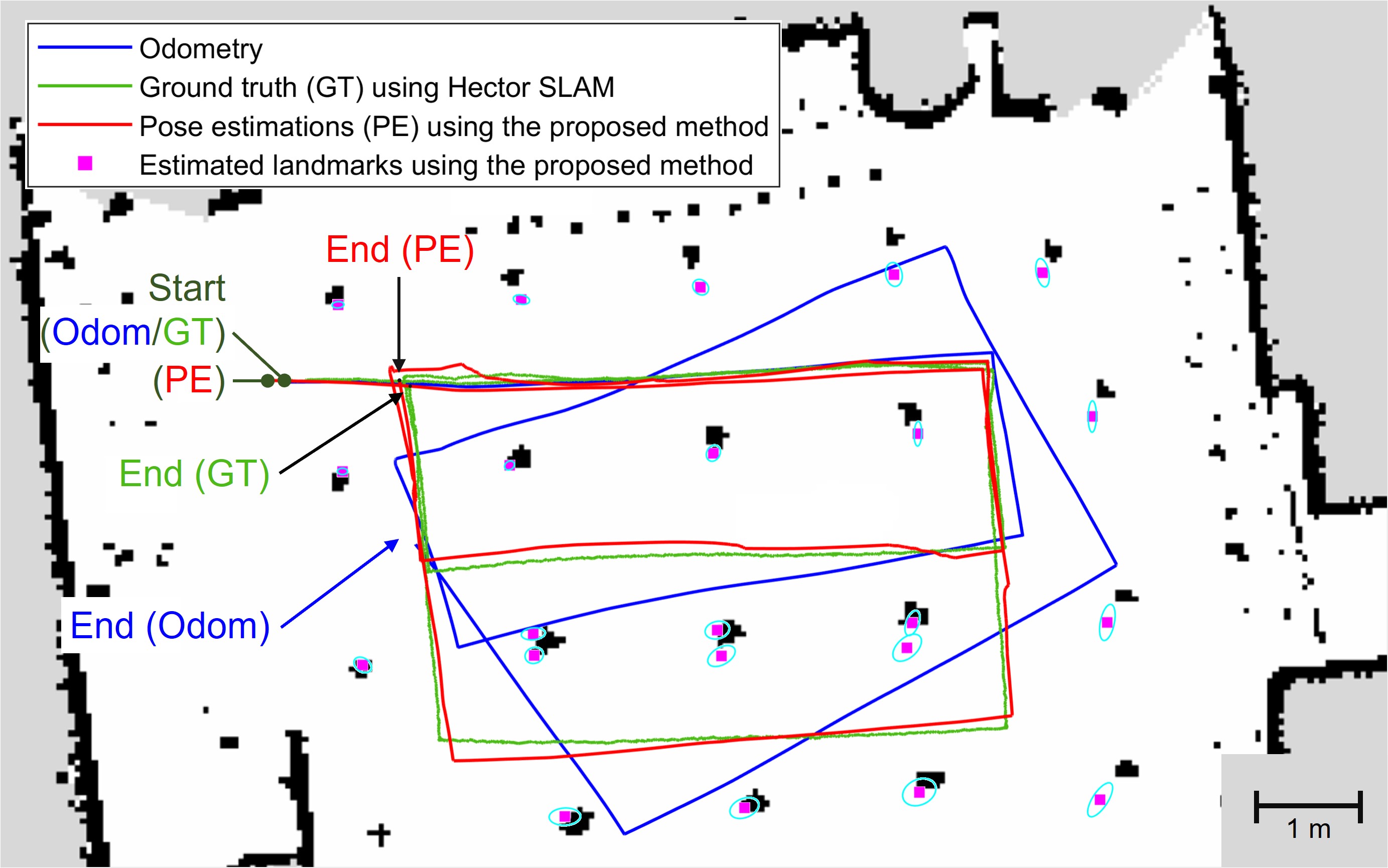}}
	\caption{ \textcolor{black}{Experiment 2 in the classroom - The estimated robot poses and the map constructed by the proposed UWB radar SLAM system. The error ellipses around the landmarks illustrate uncertainties.}}
	\label{exp2}
\end{figure}

The second experiment was conducted in a classroom where study tables could be detected as landmarks. The carpeted floor tends to drift the odometry more than in the lab floor. Moreover, the landmarks were somewhat distant from each other ($>$ 1 m apart). Hence, Both motion uncertainty $R$ and the Mahalanobis distance threshold \textcolor{black}{$(\alpha_2 = 1.6)$} had to be increased compared to the first experiment \textcolor{black}{($\alpha_1 = 0.6 < 1.6$)}. However, the results were quite satisfactory (see Fig. \ref{exp2}). Similar to the first experiment, the radar SLAM initially tends to follow odometry and after detecting a previously observed landmark, it corrects its pose.  \textcolor{black}{Compared to the ground truth, the estimated poses had a ATE of 62~mm.}

Notice that in some instances, multiple landmarks are assigned in close proximity despite being belonged to the same landmark in the ground truth (see Fig. \ref{exp2}). This is one of the issues in SLAM with unknown correspondences. This is worse when it comes to high motion uncertainty due to \textcolor{black}{noisy} odometry. It can be suggested to increase $\alpha$, but there will discrepancy with nearby landmarks after a certain limit. Hence, it is of interest to implement a suitable map management or a feature matching methodology to identify previously visited landmarks in the future.

\section{Conclusion}\label{conc}

This paper presented an anchorless approach using UWB radar for SLAM in an unknown environment. XeThru X4M300 IR-UWB sensors were used to obtain radar observations. Instead of the default downconverted baseband output with large range samples, filtered raw observations were used with small range samples to improve spatial accuracy. The proposed method \textcolor{black}{trilaterates} two radar observations to identify the point landmarks in the environment. DBSCAN-based algorithm is then proposed for further filtering before applying EKF SLAM with unknown correspondences.

\textcolor{black}{Experiments were carried out in a lab environment and a classroom in order to validate the proposed method.}  \textcolor{black}{The results show that the robot has been able to mitigate the odometry drift with sufficient accuracy while reconstructing the map of the environment successfully.} If higher accuracy is desired, a radar sensor array can be used to improve the observation accuracy. On the other hand, currently, the proposed system is incapable of classifying the feature type \textcolor{black}{(i.e. point and line features)}. Hence, we intend to develop a robust feature classification method in the future. Moreover, we plan on applying machine learning to uniquely identify landmarks using \textcolor{black}{their inherent UWB radar signatures \cite{RN144}}.


%

\ifCLASSOPTIONcaptionsoff
  \newpage
\fi

\bibliographystyle{IEEEtran}
\bibliography{mybib}

\end{document}